\documentclass{article}
\usepackage{amssymb}
\usepackage{graphicx} 
\usepackage{subcaption}
\usepackage{booktabs}
\usepackage{amsmath}
\usepackage{mathtools}
\usepackage{amsthm}
\usepackage{wrapfig}
\usepackage{enumitem}

\newtheorem{proposition}{Proposition}[section]
\usepackage{multirow,tabularx,booktabs}

\usepackage{bm}
\usepackage{siunitx}
\usepackage[table]{xcolor}

\usepackage[final]{corl_2025} 

\title{Shortcut Learning in Generalist Robot Policies: The Role of Dataset Diversity and Fragmentation}

\author{
  Youguang Xing$^{1*}$, Xu Luo$^{1*}$, Junlin Xie$^{1}$, Lianli Gao$^{1}$, Hengtao Shen$^{2}$, Jingkuan Song$^{2}$$^{\dagger}$\\
  $^{1}$UESTC, $^{2}$Tongji University\\\texttt{ygxing@std.uestc.edu.cn, frank.luox@outlook.com}
}

\begin{document}
\maketitle


\renewcommand*{\thefootnote}{\fnsymbol{footnote}}
\footnotetext{
$^\ast$: denotes equal contribution. $^\dag$: denotes corresponding author.
}

\renewcommand*{\thefootnote}{\arabic{footnote}}

\begin{abstract}
    Generalist robot policies trained on large-scale datasets such as Open X-Embodiment (OXE) demonstrate strong performance across a wide range of tasks. However, they often struggle to generalize beyond the distribution of their training data. In this paper, we investigate the underlying cause of this limited generalization capability.  We identify \emph{shortcut learning}—the reliance on task-irrelevant features—as a key impediment to generalization. Through comprehensive theoretical and empirical analysis, we uncover two primary contributors to shortcut learning: (1) limited diversity within individual sub-datasets, and (2) significant distributional disparities across sub-datasets, leading to dataset fragmentation. These issues arise from the inherent structure of large-scale datasets like OXE, which are typically composed of multiple sub-datasets collected independently across varied environments and embodiments. Our findings provide critical insights into dataset collection strategies that can reduce shortcut learning and enhance the generalization ability of generalist robot policies. Moreover, in scenarios where acquiring new large-scale data is impractical, we demonstrate that carefully selected robotic data augmentation strategies can effectively reduce shortcut learning in existing offline datasets, thereby improving generalization capabilities of generalist robot policies, e.g., $\pi_0$, in both simulation and real-world environments. More information at our website\footnote{Poject page: {\url{https://lucky-light-sun.github.io/proj/shortcut-learning-in-grps/}}}.

\end{abstract}



\section{Introduction}
    The recent advancements in machine learning, particularly in domains such as computer vision and natural language processing, can be largely attributed to the scaling up of both data and model sizes. Notably, scaling laws in these domains \cite{kaplan2020scaling, hoffmann2022training, zhai2022scaling} indicate a consistent trend of performance improvement and emergent generalization capabilities as the number of model parameters and the volume of data are increased.

    \begin{figure*}[t]
      \centering
      \includegraphics[width = 1.0\linewidth]{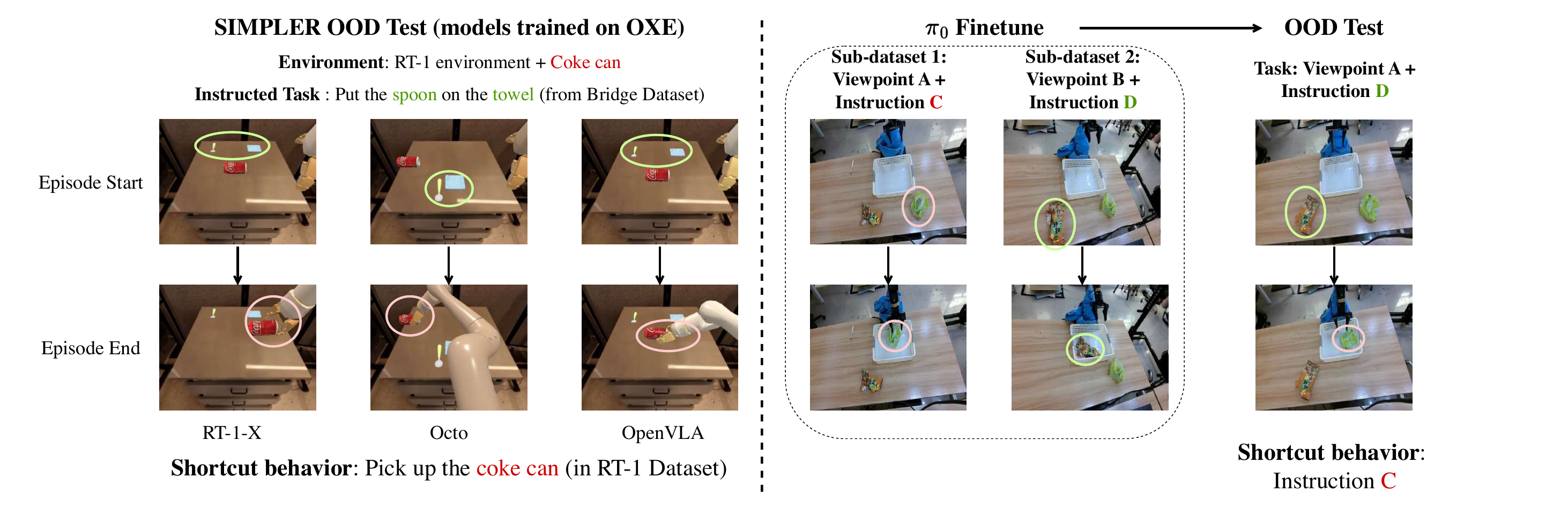}

      \caption{\textbf{Demonstrations of shortcut learning in generalist robot policies.} \textbf{Left:} Three generalist robot policies trained on the OXE dataset exhibit shortcut behavior in the SIMPLER environment \cite{Simpler}. Despite being tasked with ``put the spoon on the towel'', a task present in the Bridge sub-dataset \cite{bridge}, all models consistently perform the task ``pick up the coke'' which is exclusive to the RT-1 sub-dataset \cite{rt-1}. \textbf{Right: }$\pi_0$ \cite{pi_0} policy after finetuning on real-world data exhibits shortcut behavior. The policy was finetuned on two distinct data subsets: (Viewpoint A, Instruction C) and (Viewpoint B, Instruction D). When tasked with instruction D from the novel configuration of Viewpoint A, the policy incorrectly executes Instruction C. This indicates that the policy has learned to associate the viewpoint with the action, ignoring the provided instruction.}
      

      \label{fig:intro}
    \vspace{-0.5cm}
    \end{figure*}

    
    It is anticipated that analogous trends will emerge in the field of robotics. Consequently, recent research efforts in the field of robot learning have concentrated on the development of increasingly large-scale robot datasets \cite{rt-1,bridge,roboagent,droid,oxe,bu2025agibot} and the training of high-capacity models \cite{rt-1,rt-2,octo,openvla,rdt,pi_0,fast,team2025gemini} on these datasets, which directly map observations to actions, e.g., Vision-Language-Action (VLA) models \cite{rt-2}. The hope is that, by feeding abundant web and robot data, we can develop a generalist robot policy capable of addressing a wide spectrum of tasks and, more importantly, generalizing to novel tasks and environments out of box.

    Despite advancements in training models on large-scale datasets like Open X-Embodiment (OXE) \cite{oxe}, these models continue to demonstrate limited generalization capabilities across multiple axes, including visual, semantic, and behavioral aspects \cite{gao2025taxonomy}. This limitation cannot be ascribed to a deficiency in data, as the scale of OXE—comprising over one million episodes—surpasses that of datasets typically employed for training vision-language models, which generally consist of fewer than one million images and yet exhibit strong generalization capabilities \cite{llava}. \emph{So, what hinders generalization in robot policies?}
    


In this paper, we identify \emph{shortcut learning}---a model's reliance on spurious correlations between actions and task-irrelevant components of observations---as a significant contributor to this limitation in generalization. As illustrated in Figure \ref{fig:intro}, by learning from confounding factors such as viewpoint, background, and texture, the model fails to capture the true causal relationships between observations and actions. Consequently, it may overlook essential elements like language instructions and target objects, thereby restricting its ability to generalize beyond the training distribution.
    

    To investigate the root causes of shortcut learning in generalist robot policies, we conduct a detailed analysis of the widely used OXE dataset. Our visual and textual feature analysis reveals two critical issues: (1) limited diversity within sub-datasets, and (2) significant disparities between them, resulting in dataset \emph{fragmentation}.  Through theoretical analysis and controlled experiments, we demonstrate that both characteristics contribute to shortcut learning. Based on these findings, we derive key insights for improved robot dataset collection strategies, summarized below:
\begin{enumerate}[leftmargin=*]
    \item Ensure diversity in both task-relevant and task-irrelevant observation factors within each sub-dataset (Figure \ref{fig:LIBERO_exp}), while maintaining factor independence during data collection (Figure \ref{fig:10_subdatasets}).
    \item Maintain substantial overlap in the most important factors across sub-datasets (Figure \ref{fig:LIBERO_exp}), preserving consistency for less critical factors (Section \ref{sec:Discussion}). 
    \item Allow slightly larger distributional disparities for task-relevant factors between sub-datasets, while minimizing disparities in task-irrelevant factors (The last paragraph in Section \ref{sec:theoretical_analysis}).
\end{enumerate}


    Furthermore, we give suggestions on how to alleviate the shortcut learning in existing offline datasets, facilitating scenarios where collecting new data is infeasible. We demonstrate that carefully selected robotic data augmentation strategies can effectively increase the diversity within sub-datasets and decrease their differences. Our experiments, conducted in  SIMPLER \cite{Simpler} and real-world environment, confirm that these augmentation strategies can significantly alleviate shortcut learning in Generalist Robot Policies like $\pi_0$, and improve generalization performance.

\section{Analysis of Dataset Diversity and Fragmentation of Robot Datasets}
\label{sec:Dataset_Analysis}
In this section, before delving into the details of shortcut learning, we analyze the sub-dataset diversity and fragmentation of current large-scale robot datasets. Our discussion centers on the OXE dataset \cite{oxe}, the largest open-source robot dataset utilized for the pretraining of generalist robot policies \cite{oxe,octo,openvla,pi_0,fast,gr00t}. Specifically, we focus on OXE Magic Soup++ \cite{octo}, which comprises 27 sub-datasets from OXE that have been carefully selected to ensure high quality and have been used in several models \cite{octo,openvla,pi_0,fast}. Given that current robot datasets for large-scale pretraining similarly consist of diverse sub-datasets \cite{rdt,robomind,reed2022generalist,gr00t}, the insights derived in this section are broadly applicable.



As most generalist robot policies use vision observations and language instructions for making actions, we utilize visual and language features of the datasets to measure the diversity within sub-datasets and disparities between them, as also suggested by \cite{gao2025taxonomy}. For visual features, we use the concatenation of features from pretrained DINOv2 \cite{dinov2} and SigLIP \cite{SigLip}  as they are shown to give complementary information about images \cite{tong2024eyes,karamcheti2024prismatic}. We focus solely on the initial visual observation of each episode, as subsequent frames typically exhibit minimal variation. Language features are extracted using CLIP~\cite{CLIP} for its strong vision-language alignment. To quantify the diversity within a sub-dataset $D_i$, we adopt the uniformity metric proposed by~\cite{wang2020understanding}:
\[S_{\mathrm{diversity}}^{D_i}\triangleq \frac{1}{\mathbb{E}_{u,v\sim D_i}\big[e^{-t\|u-v\|^2_2}\big]},\]
which is maximized when the feature vectors within the sub-dataset are uniformly distributed on the unit sphere~\cite{wang2020understanding}, indicating maximum diversity (we normalize all features before calculation). This aligns with the entropy measure used in Section~\ref{sec:Theoretical_and_Empirical_Analysis} to quantify diversity. As illustrated in Figure~\ref{fig:metric_function}, the temperature parameter $t$ serves as a soft threshold, modulating the influence of pairwise distances $\|u-v\|_2^2$ on the diversity metric. A larger $t$ sharpens the effective influence range, ensuring only highly similar vectors contribute significantly.

Similarly, the disparity metric is defined as the inverse of the expected pairwise similarity between datasets:
\[S_{\mathrm{disparity}}\triangleq\frac{m(m-1)}{\sum_{i\neq j}\mathbb{E}_{u\sim D_i,v\sim D_j}\big[e^{-t\|u-v\|^2_2}\big]},\]
where we assume there are $m$ sub-datasets. In the followings, we derive insights by comparing \( S_{\mathrm{diversity}} \) and \( S_{\mathrm{disparity}} \) metrics obtained from the OXE dataset with those from commonly used vision and multimodal datasets for pretraining large-scale vision models and vision-language models.

\begin{figure*}[t]
      \centering
      \includegraphics[width = 1.0\linewidth]{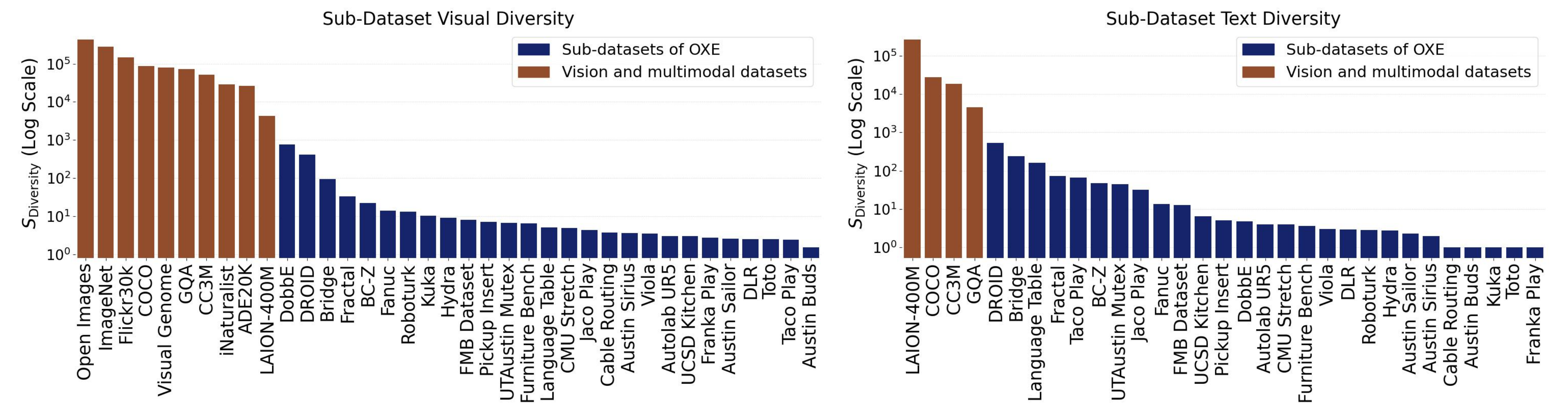}
      \caption{\small{\textbf{Comparison of visual (left) and text (right) diversity (log scale) between OXE Sub-Datasets and vision/multimodal Datasets.} OXE sub-datasets exhibit significantly lower diversity compared to their vision and multimodal counterparts. We simply chose $t=20$ as it does not influence the general trend.}}
      \label{fig:diversity}
      \vspace{-0.3cm}
\end{figure*}

\begin{figure*}[t]
      \centering
      \includegraphics[width = 0.8\linewidth]{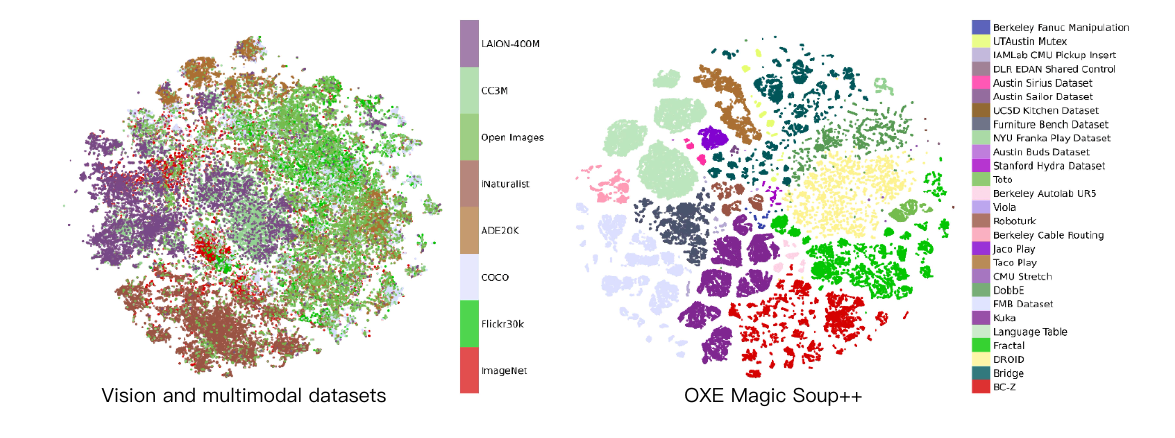}
      \caption{\small{\textbf{Comparison of t-SNE visualizations for vision/multimodal datasets (left) versus OXE Magic Soup++ (right).} The figure shows the clear data fragmentation in the OXE dataset, in contrast to the more intertwined data structure observed in the visual and multimodal datasets.}}
      \label{fig:tsne}
      \vspace{-0.3cm}
\end{figure*}
\textbf{Large-scale robot datasets exhibit limited diversity within individual sub-datasets.} As depicted in Figure \ref{fig:diversity}, the visual and textual diversity across all sub-datasets within OXE is markedly lower than that of vision and multimodal datasets. Even the most recent dataset, DROID \cite{droid}, which aims to improve diversity, remains significantly less diverse by several orders of magnitude. This limited diversity within sub-datasets primarily stems from intrinsic constraints in the collection process of robot datasets. Factors such as scenes and viewpoints are challenging to vary significantly across episodes, resulting in a lack of portability compared to web-based vision-language datasets. Another reason is that, as shown in Figure \ref{fig:oxe_task_diversity}, the robotic skills within each sub-dataset are typically predefined, restricting them to a narrow spectrum of tasks.

\begin{wrapfigure}{r}{0.3\textwidth}

\vspace{-0.7cm}
\begin{center}
\includegraphics[width=0.99\linewidth]{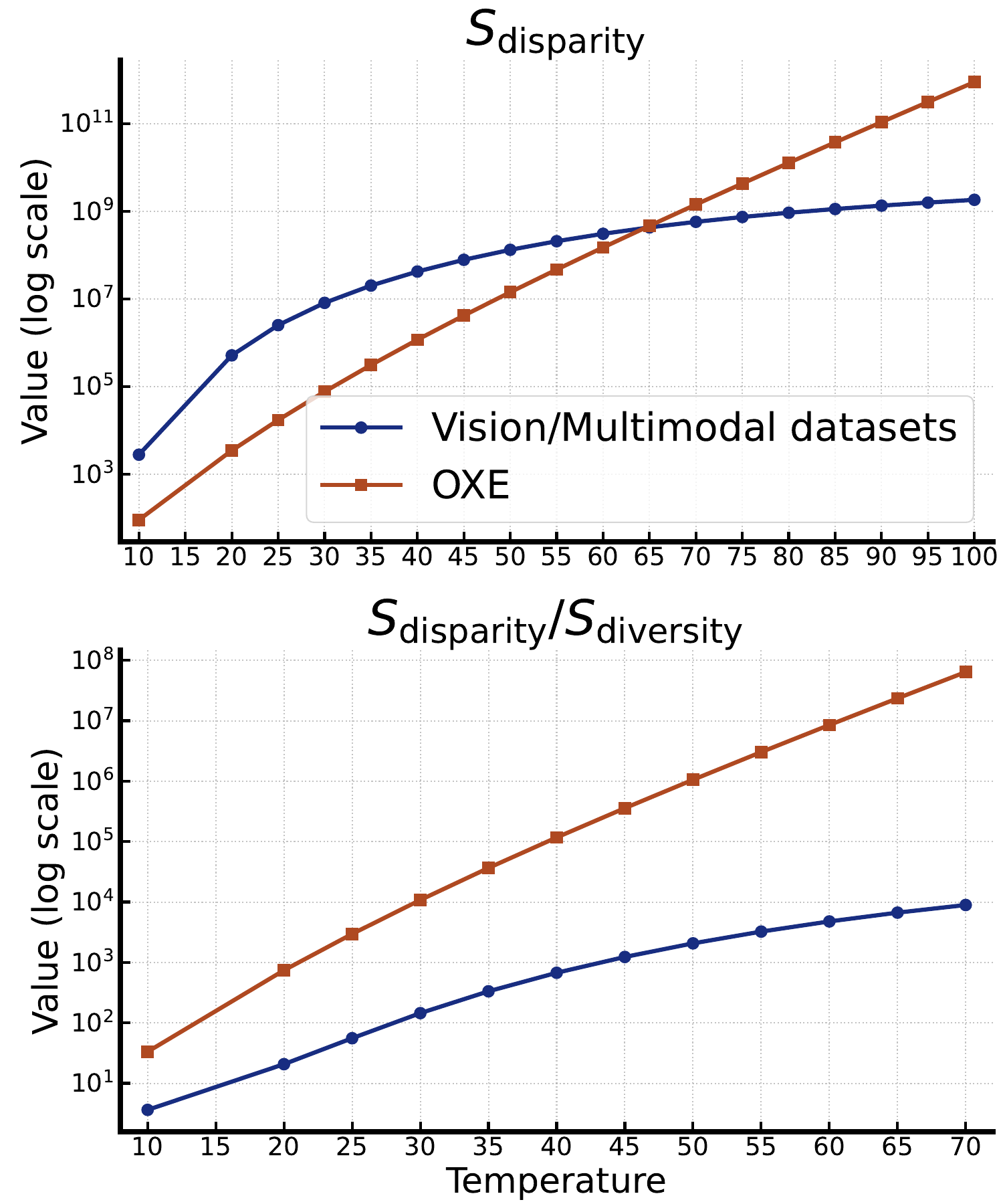}
\vspace{-0.8cm}
\end{center}
\caption{\small{Comparison of the visual disparity metric $S_{\mathrm{disparity}}$ (top) and the combined metric $\frac{S_\mathrm{disparity}}{S_\mathrm{diversity}}$ (bottom) between OXE and vision/multimodal datasets at different temperatures.}}
\vspace{-0.6cm}
\label{fig:disparity}
\end{wrapfigure}
\textbf{Large-scale robot datasets are fragmented across sub-datasets.} We present the visualization of visual features using t-SNE in Figure \ref{fig:tsne}. Unlike vision and multimodal datasets, where different datasets are often intertwined, the sub-datasets of OXE exhibit distinct separations with minimal overlap. Furthermore, some sub-datasets have several separated clusters, effectively fragmenting the whole dataset into more sub-datasets with smaller size. We show examples in Appendix \ref{sec:Appendix_Dataset_Analysis} and further discuss this in Section \ref{sec:experimental_verification}.

The top plot in Figure \ref{fig:disparity} presents the disparity metric $S_\mathrm{disparity}$ for OXE. Notably, it is higher than that of vision/multimodal datasets at higher temperatures and lower at lower temperatures. This characteristic is typical of robot datasets: distances between data points from different OXE sub-datasets are concentrated within a specific range. Conversely, distances  between data points from different vision and multimodal datasets are more dispersed. Two key factors contribute to this pattern: (1) Robotic scenarios are usually confined to limited in-room tabletop domains, which restricts the maximum possible distances and results in overall high similarity, thus lower disparity at lower temperatures.
(2) Fragmentation of data across sub-datasets prevents distances from falling below a certain threshold, establishing a lower bound, which leads to higher disparity when only close data points are considered at higher temperatures.


The bottom plot in Figure \ref{fig:disparity} illustrates the ratio $\frac{S_\mathrm{disparity}}{S_\mathrm{diversity}}$, which integrates both diversity and disparity metrics to assess the extent of dataset fragmentation. Both a deficiency in diversity within sub-datasets  and an increase in disparity between them result in sub-datasets behaving like isolated ``points'' scattered across the space, rather than forming a cohesive, interconnected dataset.

\textbf{Task instructions are distinct but similar across sub-datasets.} Despite the lack of overlap in task instructions between sub-datasets, as shown in the left plot of Figure \ref{fig:text_disparity} and Table \ref{tab:Text framentation}, text features from different sub-datasets are closer in space compared to those from vision and multimodal datasets. This similarity arises from shared robotic skills, such as pick-and-place and open/close tasks, and the consistency of text instructions within the same domain.






\section{The Role of Dataset Diversity  and Fragmentation in Shortcut Learning}
\label{sec:Theoretical_and_Empirical_Analysis}
In this section, we prove that both of the lack of diversity within sub-datasets and a large disparity (fragmentation) between them lead to shortcut learning. We first describe shortcut learning in detail.
\subsection{Shortcut Learning}
In the standard supervised learning or imitation learning framework, we consider the observation and target random variables, denoted as $x$ and $y$, respectively, with the conditional probability distribution $p(y|x)$. The objective is to learn this distribution using a model $\pi_\theta(y|x)$, based on a training dataset $\{x_i, y_i\}_{i=1}^N$ sampled from the training distribution, where $x \sim p_{\mathrm{train}}(x)$ and $y \sim p(y|x)$. Following the approach in \cite{darla}, we assume that the observation $x$ is generated by multiple ``observation factors'' $[c_1, c_2, \ldots, c_d] \in \mathcal{C}^d \subset{\mathbb{R}}^d$. For instance, in manipulation tasks, these factors may include image background, object type, object position, robot arm type, language instructions, among others. We define $u \subset \mathcal{C}^d$ as the set of task-relevant factors that causally influence the label $y$, and $v \subset \mathcal{C}^d$ as the set of all other task-irrelevant factors. Formally, the ground-truth probability satisfies $p(y|x=g(u,v)]) = p(y|u)$ and $p(y, v) = p(y)p(v)$, where $g$ is the factor generation model.

Shortcut learning is characterized by the scenario where the learned model $\pi_\theta$ relies on the irrelevant factors $v$ or equivalently, $\pi_\theta(y|x) \neq \pi_\theta(y|u)$. \textbf{Shortcut learning arises when, within the training distribution, the task-relevant factors $u$ and task-irrelevant factors $v$ are not independent, i.e., $p_{\mathrm{train}}(u, v) \neq p_{\mathrm{train}}(u)p_{\mathrm{train}}(v)$.} Due to the causal relationship between $u$ and $y$, $v$ and $y$ become correlated, leading to spurious correlations between $v$ and $y$. Consequently, the model $\pi_\theta$ may inadvertently learn from these spurious correlations, resulting in shortcut learning of factors in $v$, which adversely affects the model's performance on out-of-distribution data. In the subsequent sections, we assess the correlation between $u$ and $v$ to quantify the extent of the shortcut.

\subsection{The Reasons Behind Shortcut Learning on Robot Data}
\label{sec:theoretical_analysis}


We first establish a mathematical framework to analyze how correlations can arise in a dataset composed of multiple distinct sub-datasets. Consider a dataset $D$ characterized by two random variables, $u \sim p_u(u)$ and $v \sim p_v(v)$, with supports $U, V \subset \mathbb{R}$. We model $D$ as a mixture of $m$ sub-datasets, $\{D_1, D_2, \ldots, D_m\}$, where each sub-dataset $D_i$ has its own distributions $u_i \sim p_{u_i}(u_i)$ and $v_i \sim p_{v_i}(v_i)$ with supports $U_i$ and $V_i$. The overall supports are thus $U = \cup_{i=1}^m U_i$ and $V = \cup_{i=1}^m V_i$.

We make the following simplifying assumptions for our analysis:
\begin{enumerate}
    \item Intra-dataset Independence: Within any given sub-dataset $D_i$, the variables $u_i$ and $v_i$ are independent, i.e., $p_i(u, v) = p_{u_i}(u) p_{v_i}(v)$.
    \item Uniform Mixture: The overall dataset $D$ is a uniform mixture of the sub-datasets, such that $p_u(u) = \frac{1}{m} \sum_{i=1}^m p_{u_i}(u)$ and $p_v(v) = \frac{1}{m} \sum_{i=1}^m p_{v_i}(v)$.
\end{enumerate}
The first assumption is approximately valid, as each sub-dataset is collected
under controlled conditions, minimizing the introduction of dependencies between factors. 
To quantify the correlation between $u$ and $v$ across the entire dataset $D$, we use the normalized mutual information:
\[
\overline{I}(u,v)=\frac{2I(u,v)}{H(u)+H(v)},
\]
where $I(u,v)$ is the standard mutual information and $H(\cdot)$ is the Shannon entropy. For simplicity, the following propositions are presented for the case of $m=2$ sub-datasets.

\begin{proposition}[Mutual Information in Disjoint Sets]
\label{prop:diversity}
Given two sub-datasets where the supports for both variables are disjoint, i.e., $U_1\cap U_2=\varnothing$ and $V_1\cap V_2=\varnothing$, the normalized mutual information between $u$ and $v$ is given by:
\begin{equation}
    \overline{I}(u,v)=\frac{4}{C_{\mathrm{diversity}}+4},
\end{equation}
where $C_{\mathrm{diversity}} = H(u_1)+H(u_2)+H(v_1)+H(v_2)$ is the sum of the entropies within each sub-dataset.
\end{proposition}

\begin{proposition}[Mutual Information in Overlapping Sets]
\label{prop:fragmentation}
Given two sub-datasets with potentially overlapping supports, let $U_{12}=U_1\cap U_2$ and $V_{12}=V_1\cap V_2$. The normalized mutual information is bounded by:
\begin{equation}
    \overline{I}(u,v)\leq 1-\frac{C_{\mathrm{diversity}}}{C_{\mathrm{diversity}}+(4-C_{\mathrm{interleave}})},
\end{equation}
where $C_{\mathrm{interleave}}=\sum_{u\in U_{12}}\left[p_{u_1}(u)+p_{u_2}(u)\right] + \sum_{v\in V_{12}} \left[p_{v_1}(v)+p_{v_2}(v)\right]$ quantifies the degree of overlap (interleaving) between the sub-datasets.
\end{proposition}

We hypothesize that the mathematical framework presented above explains how shortcut learning emerges from the structural properties of large-scale robot datasets. Our central hypothesis is that the spurious correlations exploited by models are a direct consequence of dataset fragmentation and a lack of intra-dataset diversity. We bridge the mathematical propositions to this real-world phenomenon with the following assumptions:
\begin{itemize}
    \item We model the \emph{task-relevant factors} (e.g., object positions, language instructions) and \emph{task-irrelevant factors} (e.g., viewpoints, backgrounds) in a robot dataset as the variables $u$ and $v$, respectively.
    \item We assume that large datasets like OXE, which are composed of many independently collected sub-datasets, can be approximated by our mixture model. The low diversity within these sub-datasets and the significant disparities between them correspond to the low-entropy and disjoint/low-overlap conditions of our propositions.
\end{itemize}

Under this hypothesis, the mathematical results provide a formal basis for our core claims:

\textbf{Lack of diversity strengthens spurious correlations.} Proposition \ref{prop:diversity} mathematically demonstrates this intuition. It shows that when sub-datasets are highly fragmented (disjoint supports), the mutual information (our proxy for spurious correlation) is inversely proportional to $C_{\mathrm{diversity}}$ (our proxy for the total diversity within sub-datasets). A robotic model trained on such a dataset can easily learn to associate a task-irrelevant factor (e.g., a specific viewpoint) with a particular sub-dataset, which in turn reveals information about the task-relevant factor, creating a shortcut.

\textbf{Interleaving sub-datasets weakens spurious correlations.} Proposition \ref{prop:fragmentation} provides theoretical support for this claim. It shows that as the degree of interleaving ($C_{\mathrm{interleave}}$) increases, the upper bound on the mutual information tightens and moves towards zero. Intuitively, when sub-datasets share common factors (e.g., the same objects appear from multiple viewpoints), it becomes impossible for the model to use those factors as reliable shortcuts to identify the sub-dataset of origin, forcing it to learn the true causal relationships.

\textbf{Impact of the disparity between non-overlapping sub-datasets on shortcut learning.} While Proposition~\ref{prop:fragmentation} elucidates how sub-dataset intersections affect shortcut learning, the influence of distance between non-overlapping sub-datasets remains less clear. As shown in Section~\ref{sec:Dataset_Analysis}, the OXE dataset exhibits minimal interleaving of visual and textual features across sub-datasets, yet the textual feature distance ($u$) is notably smaller than the visual distance ($v$). We hypothesize that larger sub-dataset distances in task-irrelevant features exacerbate shortcut learning. This stems from two key observations: (1) neural networks prioritize learning simpler patterns first~\cite{memorization,spectral}, and (2) larger feature distances imply greater variance. When task-irrelevant features have substantially greater between-sub-dataset distances than task-relevant ones, models preferentially learn these higher-variance features, forming shortcuts. In OXE, this explains the model's tendency to rely on visual cues over text instructions (Figure~\ref{fig:intro}). We formalize this intuition through gradient analysis of linear models in Appendix~\ref{sec:linear_model_analysis}.

\subsection{Experimental Verification on LIBERO}
\label{sec:experimental_verification}

In this section, we validate the conclusions drawn in Section \ref{sec:theoretical_analysis} through controlled experiments using the LIBERO benchmark \cite{libero}, which features a Franka Emika Panda arm in simulation with demonstrations containing camera images, language instructions, and delta end-effector pose actions.

\textbf{Task Setup.}
To empirically validate our theoretical claims that low intra-dataset diversity and high inter-dataset disparity foster shortcut learning, we conduct controlled experiments on the LIBERO-Spatial task suite. In this setup, we define the task-relevant factors ($u$) as the object's position and the corresponding language instruction. The camera viewpoint serves as the primary task-irrelevant factor ($v$), mirroring the significant viewpoint variations observed across sub-datasets in large-scale robot datasets like OXE.

\begin{wrapfigure}{r}{0.3\textwidth}

\vspace{-0.7cm}
\begin{center}
\includegraphics[width=0.99\linewidth]{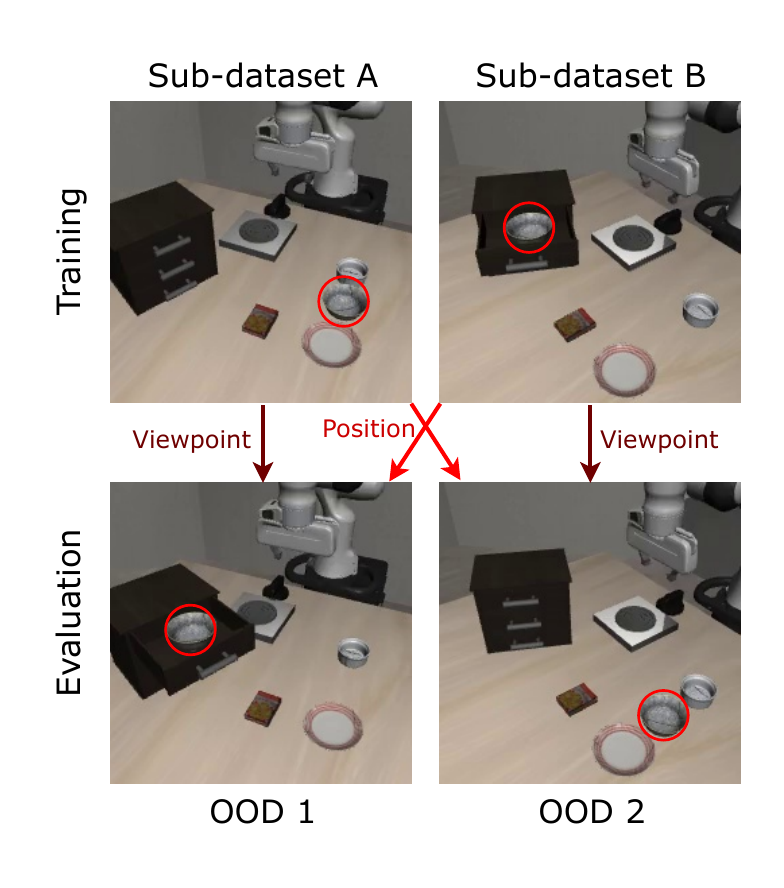}
\vspace{-0.8cm}
\end{center}
    \caption{\small{An example of our LIBERO experiment setting, with only one task (or equivalently, one object position/language) within each sub-dataset.}}
\vspace{-0.8cm}
\label{fig:environment_setting}
\end{wrapfigure}

\textbf{Training and Evaluation Protocol.}
For each experiment, we construct a training dataset composed of two distinct sub-datasets, $D_A$ and $D_B$. Each sub-dataset is generated to create a strong spurious correlation between the task-irrelevant viewpoint and the task-relevant position. For instance, demonstrations in $D_A$ exclusively pair a specific range of viewpoints (Viewpoint Range A) with a specific set of object positions (Position Set A), while $D_B$ pairs Viewpoint Range B with Position Set B. The model is trained on the combination of $D_A$ and $D_B$. To quantify shortcut learning, we evaluate the trained policy on out-of-distribution (OOD) configurations where the learned spurious correlations are broken. Specifically, the evaluation consists of two controlled settings: (1) tasking the model with object positions from Set B but from viewpoints within Range A, and (2) the reverse pairing (positions from Set A, viewpoints from Range B). A model relying on the viewpoint shortcut would fail, as it would incorrectly associate the viewpoint with the training-time positions, ignoring the actual object position and instruction.

\textbf{Experimental Variables and Metrics.}
We systematically vary the properties of the training data to analyze their impact.
\textit{Viewpoint diversity} is the radius of the viewpoint range within each sub-dataset, while \textit{viewpoint disparity} is the distance between the centers of the two viewpoint ranges. To study the effect of task-relevant diversity and disparity, we vary the number of object positions per sub-dataset (from 1 to 5) and their spatial layout (intertwined vs. separated, see Figure~\ref{fig:Disparity_of_ten_subdataset}). Performance is measured by two key metrics: (1) the \textbf{OOD success rate}, averaged over the two OOD settings, which directly measures generalization, and (2) the \textbf{degree of shortcut learning}, a human-assessed score quantifying the model's tendency to perform the wrong task based on the irrelevant viewpoint cue (lower is better). To ensure fair comparisons, for a given model, the OOD evaluation viewpoint is kept consistent within each experimental set (e.g., a curve in one plot), where we vary data diversity or disparity. Details are available in Appendix D.1.

\begin{figure*}[t]
      \centering
      \includegraphics[width = 1.0\linewidth]{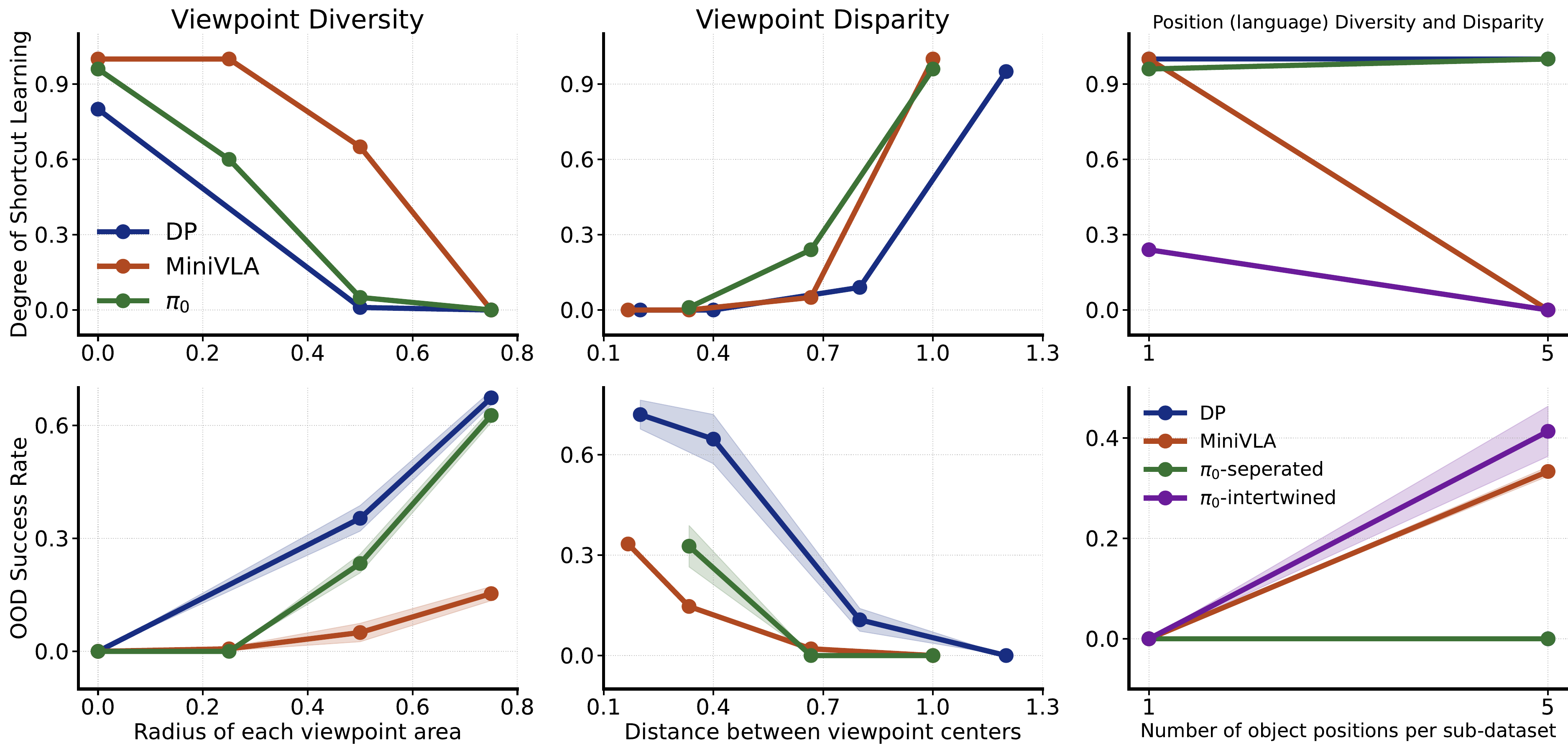}
      \caption{Impact of sub-dataset diversity and disparity on the degree of shortcut leaning and out-of-distribution (OOD) performance of robot policies, analyzing task-relevant factors (object position, language) and task-irrelevant factors (viewpoint). \textbf{Note}: Performance metrics are not directly comparable across models due to intentionally varied experimental settings (see Appendix \ref{sec:Experiment_Details}).}
      \label{fig:LIBERO_exp}
      \vspace{-0.7cm}
\end{figure*}

\textbf{Models.} We evaluate three models: (1) \textbf{Diffusion Policy} \cite{Diffusion_policy}, a purely visual policy without language input, utilizing a ResNet-18 architecture; (2) \textbf{MiniVLA} \cite{MiniVLA}, a VLA with the same autoregressive structure as OpenVLA \cite{openvla}, but with fewer than 1 billion parameters; (3) $\bm{\pi_0}$ \cite{pi_0}, a strong VLA employing a flow matching objective, pretrained on large-scale robot datasets. While Diffusion Policy and MiniVLA are trained from scratch, we finetune $\pi_0$ from the pretrained checkpoints with LoRA in order to investigate whether powerful models pretrained on extensive robot datasets are still prone to shortcut learning.

\textbf{Results.} As shown in Figure~\ref{fig:LIBERO_exp}, enhancing diversity within sub-datasets and minimizing disparity between them effectively reduces shortcut dependencies across all evaluated models, aligning with our theoretical analysis. This improvement holds for both task-irrelevant (e.g., viewpoint) and task-relevant factors (e.g., object positions and language variations). Notably, when diversity is increased or disparity decreased, all models transition from complete shortcut reliance (zero success rate) to shortcut-free performance (nonzero success rates). Through further inspection, we find that shortcut mitigation occurs before task mastery emerges—the model first abandons the shortcut (a smaller degree of shortcut learning) before learning the correct behavior necessary to achieve non-zero success rates. We also note that, increasing object position diversity does not mitigate shortcut learning in the diffusion policy, likely due to the absence of language input. This suggests that without linguistic cues, the model struggles to identify task-relevant features from visual observation alone, underscoring the importance of language instructions.

\begin{wrapfigure}{r}{0.6\textwidth}

\vspace{-0.7cm}
\begin{center}
\includegraphics[width=1.0\linewidth]{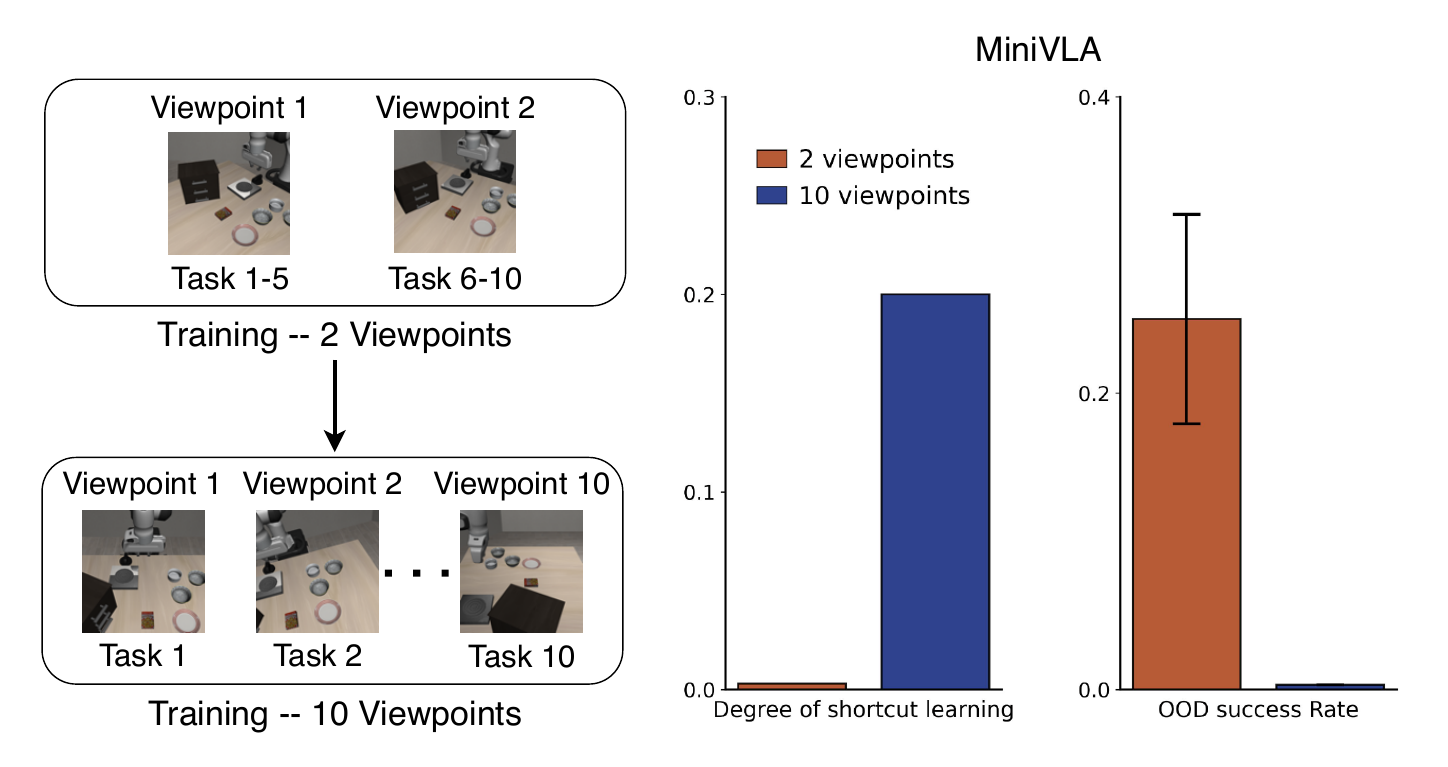}
\vspace{-0.6cm}
\end{center}
\caption{\small{\textbf{Diversity does not always help.} Increasing viewpoint diversity (2 to 10) by assigning each task a distinct viewpoint induces factor correlations in sub-datasets, aggregating fragmentation. This fosters shortcut learning and harms OOD performance.}}
\vspace{-0.45cm}
\label{fig:10_subdatasets}
\end{wrapfigure}

\textbf{Diversity does not always help.} 
Our previous analyses demonstrate that increased diversity generally mitigates shortcut learning by reducing fragmentation across sub-datasets. However, these results are obtained under the assumption of independence of factors within sub-datasets. Not all forms of diversity improve generalization. When diversity breaks factor independence within sub-datasets (e.g., some sub-datasets of OXE; see Figure \ref{fig:subdataset_fragmentation}), fragmentation worsens. As illustrated in Figure~\ref{fig:10_subdatasets}, increasing viewpoint diversity from 2 to 10—while assigning distinct viewpoints to individual tasks—introduces shortcuts and drops OOD success of MiniVLA to zero (we evaluate on the same positions and viewpoints to ensure fairness). Here, viewpoint diversity fragments the original sub-datasets into 10 disjoint subsets, exacerbating fragmentation. \textbf{This underscores the need for controlled diversity that preserves factor independence and avoids sub-dataset fragmentation during data collection.}

\subsection{Real-World Experimental Verification}

\label{sec:three objects}

To validate our theoretical conclusions from Section 3.2 in a physical environment, we conducted a real-world experiment. The setup, similar to the one depicted in Figure \ref{fig:intro}, utilized an AgileX PIPER robotic arm and two cameras positioned at different viewpoints. Initially, we constructed two distinct sub-datasets. Each sub-dataset represented a unique combination of a camera viewpoint (a task-irrelevant factor) and a target object with its corresponding instruction (task-relevant factors). As demonstrated in our preliminary findings (Figure \ref{fig:intro}), a $\pi_0$ model fine-tuned on these two highly-correlated sub-datasets exhibited severe shortcut learning; it learned to associate the viewpoint with the action, ignoring the language instruction.

\begin{wrapfigure}{r}{0.45\textwidth}

\vspace{-1.2cm}
\begin{center}
\includegraphics[width=1.0\linewidth]{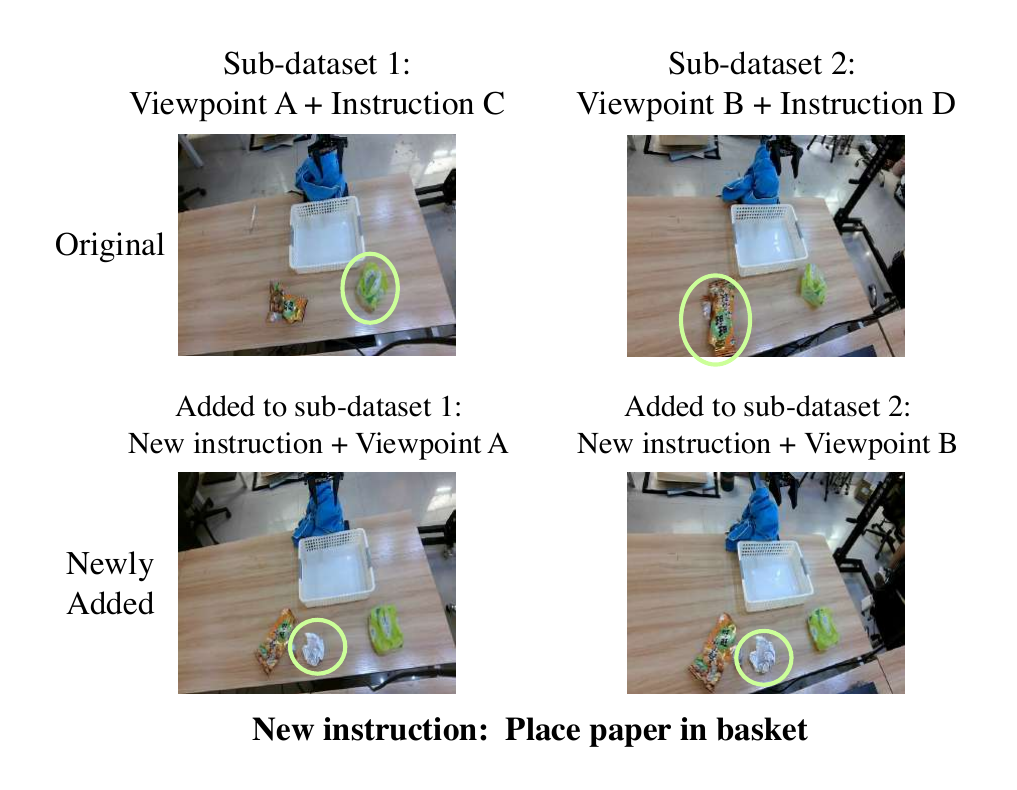}
\vspace{-0.6cm}
\end{center}
\caption{\small{Building a ``bridge'' to connect sub-datasets for the $\pi_0$ fine-tuning experiment. Data from a third object is added under both viewpoints.}}
\vspace{-0.45cm}
\label{fig:three_objects}
\end{wrapfigure}

To investigate how increasing sub-dataset diversity and reducing inter-dataset disparity could mitigate this issue, we introduced new data. Specifically, we added demonstrations involving a third target object, captured from \textit{both} camera viewpoints (as shown in the bottom row of Figure \ref{fig:three_objects}). This new data acts as a ``bridge'' between the original two sub-datasets. By doing so, we simultaneously increased the instruction diversity within each sub-dataset and decreased the disparity between them, as they now share a common instruction factor.

The results, summarized in Table \ref{tab:real-world}, are compelling. The addition of the third ``bridge'' object completely eliminated the observed shortcut behavior, leading to a substantial improvement in OOD success rate. By learning from data where the object and instruction were consistent across different viewpoints, the model successfully learned viewpoint invariance. This experiment not only confirms our theoretical framework in a real-world setting but also suggests a valuable strategy for data collection: deliberately creating ``bridge'' data by varying one factor while keeping others constant can effectively connect disparate sub-datasets, break spurious correlations, and enhance the generalization capabilities of robot policies.



\begin{table}[t]
\caption{Real-world finetuning of $\pi_0$. Introducing a third object or applying viewpoint augmentation mitigates shortcut learning and improves OOD success by increasing sub-dataset diversity and reducing disparity.}
\centering
\begin{tabular}{lcc}
\toprule
Model & Shortcut degree $\downarrow$ & OOD success rate $\uparrow$\\\midrule
$\pi_0$ baseline & 0.6 & 0.2\\
~+ third object & 0  & 0.75\\
~+ viewpoint aug & 0.15 & 0.55\\
\bottomrule
    \vspace{-0.1cm}
         
\label{tab:real-world}
\vspace{-0.35cm}
\end{tabular}
\end{table}

\section{Alleviating Shortcut Learning in Offline Datasets via Data Augmentation}

Given that collecting large-scale, perfectly balanced robot datasets from scratch is often prohibitively expensive, a practical alternative is to improve existing offline datasets. In this section, we investigate whether targeted data augmentation strategies can effectively increase sub-dataset diversity and decrease distributional disparities, thereby mitigating shortcut learning.


\subsection{Viewpoint Augmentation to Bridge Visual Gaps}

\begin{wrapfigure}{r}{0.4\textwidth}

\vspace{-2cm}
\begin{center}
\includegraphics[width=1.0\linewidth]{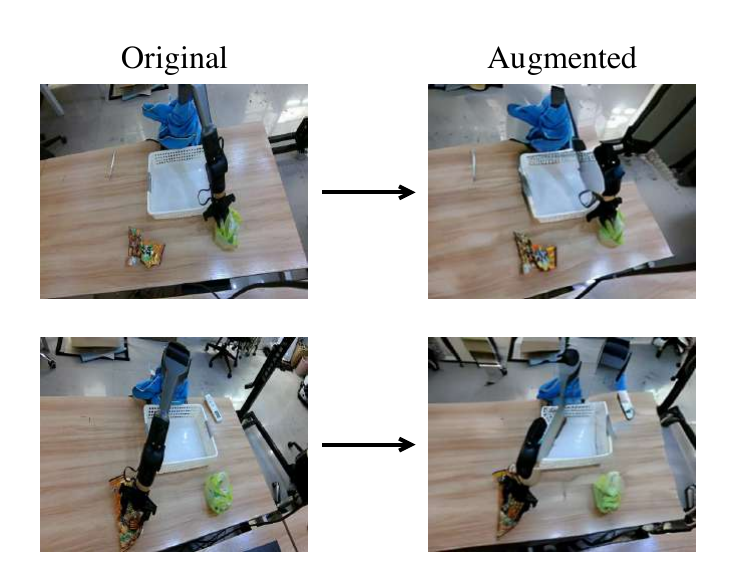}
\vspace{-0.6cm}
\end{center}
\caption{\small{Example of viewpoint augmentation. We generate novel views to create shared visual contexts between sub-datasets, breaking the spurious correlation between viewpoint and task.}}
\vspace{-0.35cm}
\label{fig:viewpoint_aug}
\end{wrapfigure}

Spurious correlations between camera viewpoints and specific tasks are a usual cause of shortcut learning. To address this, we explore the use of viewpoint augmentation methods \cite{tian2024view,chen2024rovi}. In our $\pi_0$ fine-tuning experiment (detailed in Section \ref{sec:three objects}), we synthetically expand each sub-dataset by generating images from the other's perspective. Specifically, we employ the ZeroNVS model \cite{zeronvs} to augment viewpoint A to B and vice-versa for every image, as illustrated in Figure \ref{fig:viewpoint_aug}. This process effectively breaks the fragmentation of viewpoint factors across the sub-datasets. As evidenced by the results in Table \ref{tab:real-world}, fine-tuning with viewpoint-augmented data significantly reduces the degree of shortcut learning in $\pi_0$ and leads to a higher OOD success rate.


\subsection{Object Augmentation to Unify Task Distributions}

\begin{wrapfigure}{r}{0.5\textwidth}

\vspace{-0.9cm}
\begin{center}
\includegraphics[width=0.99\linewidth]{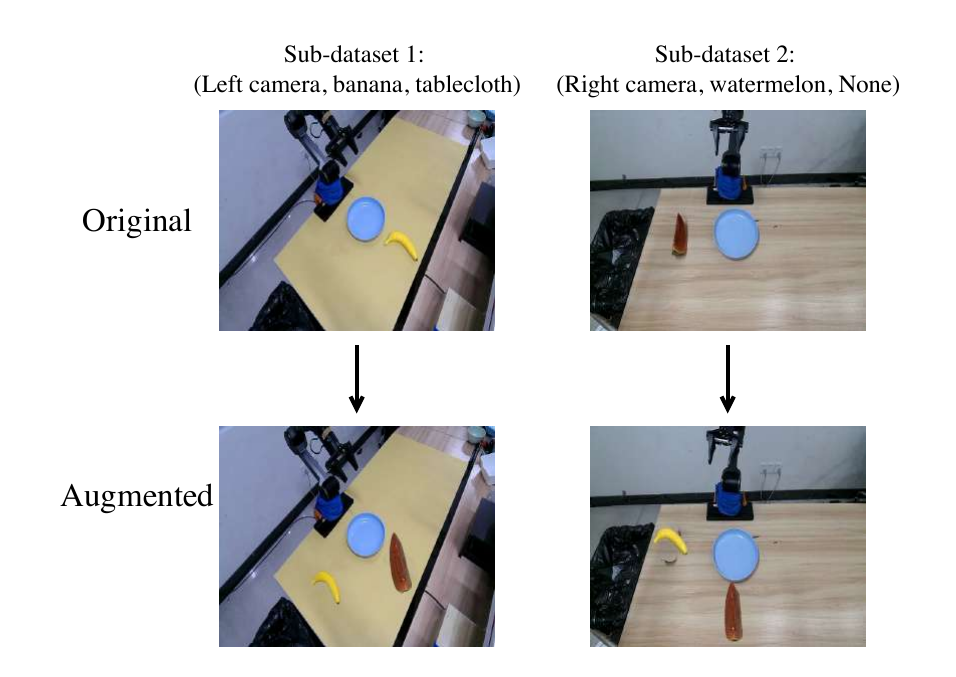}
\vspace{-0.6cm}
\end{center}
    \caption{\small{Using object augmentation to break spurious correlations in real-world data. The original images (top) show objects that are tied to specific scenes. The augmentation process (bottom) swaps these objects between the scenes, forcing the model to learn object identity independent of visual context like viewpoint or background.}}
\vspace{-0.4cm}
\label{fig:object_aug}
\end{wrapfigure}

When sub-datasets are organized around distinct target objects (as seen in Figure \ref{fig:intro}), object augmentation techniques \cite{semantically_imagined_experience,roboagent,chen2024semantically} offer a powerful solution. By programmatically swapping objects between images, we can intertwine the object and language distributions, thus reducing the disparities between sub-datasets.

We validate this approach in both the SIMPLER simulation environment \cite{Simpler}—which provides realistic versions of the Bridge and RT-1 datasets—and in real-world experiments (Figure \ref{fig:object_aug}). For the SIMPLER experiments, following the methodology of \cite{gao2025taxonomy}, we isolate the effect of real-to-sim gap by fine-tuning a pretrained $\pi_0$ model on an offline dataset collected within SIMPLER. This dataset consists of successful trajectories from pretrained models. Due to the complexity of achieving precise object segmentation and inpainting across varied conditions (e.g., embodiments, backgrounds), our analysis focuses on the reduction in shortcut behavior rather than task success rate, using a simplified implementation detailed in Appendix \ref{appendix: data_aug}. Experiment details can be found in Appendix \ref{Real-world Experiment Setup} and \ref{appendix: data_aug}.

\begin{wraptable}{r}{0.4\linewidth}
 \vspace{-0.8cm}
  \centering
  \def\arraystretch{0.9}
  \setlength{\tabcolsep}{0.42em}
\begin{tabularx}{0.95\linewidth}{ccc}
\toprule
\multirow{2}{*}{Model}&\multicolumn{2}{c}{Shortcut degree}\\\cmidrule(lr){2-3}
 & SIMPLER & Real-world\\\midrule
$\pi_0$ & 1.0 & 0.8\\
$\pi_0+\mathrm{aug}$ & 0.68  & 0.25\\
\bottomrule
    \end{tabularx}
    \vspace{-0.1cm}
         \caption{\small{Comparisons between $\pi_0$ with and without object augmentations in the SIMPLER and real-world environment.}}
\label{tab:augmentation}
\vspace{-0.4cm}
\end{wraptable}

As shown in Table~\ref{tab:augmentation}, this strategy yields significant improvements. The non-augmented $\pi_0$ model, prone to shortcut learning, completely fails to follow language instructions in OOD settings. In contrast, the augmented version exhibits substantially better language-following and object-reaching capabilities. These results confirm that carefully selected data augmentation strategies can successfully mitigate shortcut learning by enhancing sub-dataset diversity and bridging distributional gaps in existing robot datasets.

\section{Discussion and Conclusion}
\label{sec:Discussion}

Our analysis reveals that large-scale robot datasets like OXE suffer from limited sub-dataset diversity and severe fragmentation, a problem that extends even \textit{within} individual sub-datasets. This structure inherently promotes shortcut learning, meaning that simply adding more similarly-fragmented data can be detrimental to generalization.

The validity of our findings is underscored by their ability to explain the data curation strategies of recent state-of-the-art policies. For instance, some models achieve strong performance by carefully mitigating OXE's flaws: RT-X models \cite{oxe} use only 11 heavily-filtered sub-datasets, while other work \cite{remix} demonstrates that performance improves by down-weighting fragmented sub-datasets (e.g., Cable routing, Kuka) and up-sampling more cohesive ones (e.g., Toto). An even stronger testament to our conclusions comes from leading policies like $\pi_{0.5}$ \cite{pi_0.5} and Gemini Robotics \cite{team2025gemini}, which largely eschew OXE altogether. Instead, they opt for meticulously controlled datasets where specific factors—such as tasks ($\pi_{0.5}$), environments (Gemini), or embodiments (Gemini)—are deliberately fixed while others are systematically varied.

The overarching takeaway is clear: given the practical limitations of current data collection, pursuing generalization across all possible factors simultaneously is untenable. \textbf{The most effective path forward is to control the experimental setup by strategically fixing factors that are non-essential or difficult to vary, while systematically diversifying the factors of interest.} This disciplined approach is crucial for preventing shortcut learning and ensuring that models acquire robust, meaningful patterns rather than relying on spurious correlations. Ultimately, our analysis not only provides a framework for understanding the success of previous data collection strategies but also offers clear, actionable insights for optimizing the training of next-generation generalist policies.

\section{Limitations and Future Work}

While this work provides critical insights into shortcut learning in generalist robot policies, we acknowledge several limitations that open valuable avenues for future research.

\paragraph{Identifying Specific Shortcuts in Large-Scale Datasets}
Although our work demonstrates the \textit{existence} of shortcut learning, our analysis does not pinpoint the \textit{specific} spurious correlations exploited by policies trained on massive, heterogeneous datasets like OXE, nor have we investigated the hierarchy of these shortcuts. This limitation points to a clear direction for future work, which should focus on developing fine-grained diagnostic tools and interpretability methods to automatically identify the precise features that models rely on. Such research could involve causal analysis or counterfactual evaluation on large datasets to understand which shortcuts are most dominant and how they vary across different model architectures.

\paragraph{Measuring Diversity of Task-Relevant Factors}
Our quantitative analysis of dataset diversity and disparity primarily focused on task-irrelevant visual features. Due to the significant challenges of collecting and annotating large-scale behavioral data, we could not precisely measure the diversity of \textit{task-relevant} factors, such as the distribution of target object positions or grasp affordances. To address this, we encourage the development of more sophisticated metrics that can capture the complexity of action-centric and object-centric diversity. Exploring semi-supervised or self-supervised methods to automatically label these task-relevant factors would enable a more complete understanding of data quality. Exploring shortcut learning relevant to other  observation modalities like proprioceptions \cite{kuang2025adapt} and tactile observations is also a promising future direction.

\paragraph{Scalability and Generalization of Data Augmentation}
Our experiments successfully show that targeted data augmentations can mitigate shortcut learning on a controlled scale, but we have not demonstrated their effectiveness on extremely large datasets like the full OXE collection. The computational cost and potential for introducing artifacts with current augmentation models remain significant challenges at scale. Therefore, a crucial next step is to develop highly efficient, robust, and automated data augmentation pipelines suitable for millions of trajectories. Future work could also systematically compare different augmentation strategies to create a practical guide on the best trade-off between computational cost and performance gain.

\paragraph{Real-World Complexity}
Although we validated our findings in both simulation and real-world setups, the scale and complexity of our real-world experiments are inherently limited and may not fully capture all potential failure modes. Consequently, more extensive real-world studies are needed to validate our findings across a wider range of physical robots, environments, and tasks, including long-term deployments to observe if new, unforeseen shortcuts emerge over time.

\paragraph{Exploring Model-Centric Solutions}
Our proposed solutions are primarily data-centric, focusing on improving the dataset itself. We did not explore model-centric approaches, such as how different model architectures, training objectives, or regularization techniques might inherently resist shortcut learning, even when trained on fragmented data. A promising future direction is to conduct a comparative analysis of how different model architectures and learning paradigms interact with dataset biases. Investigating hybrid approaches that combine data-centric enhancements with model-centric regularization techniques could lead to the most robust and generalizable robot policies.

\bibliography{example}  

\appendix

\section{Related Work}
\label{sec:related work}
	\textbf{Generalist robot policies.} Following the trend in machine learning research, multiple works have developed robotic foundation models \cite{rt-1,rt-2,octo,rdt,doshi2024scaling,HiT,openvla,roboturk}, in particular Vision-Language-Action (VLA) models \cite{rt-2,openvla,pi_0,fast,EmbodiedCoT,navila,RT-H,team2025gemini,gr00t,pi_0.5,hirobot,Otter,MiniVLA,OpenVLA-oft}. By pretraining on increasingly large robot datasets \cite{oxe,rh20t,bridge,droid,roboagent,dexmimicgen,bu2025agibot}, these models produce generalist robot policies that excel at a wide variety of tasks and exhibit some degree of generalization \cite{rt-2,octo}. However, research by \cite{gao2025taxonomy} suggests that training on large-scale datasets does not significantly enhance the generalization capabilities of these policies. In particular, current models still struggle to generalize to many environmental changes, including viewpoint, language, object poses, etc. Our work delves into the problem and shows that limited diversity within individual sub-datasets, and significant distributional disparities across sub-datasets lead to shortcut learning of policies, which hinders generalization. Recent VLAs such as $\pi_0$ \cite{pi_0}, $\pi_{0.5}$ \cite{pi_0.5}, and Gemini Robotics \cite{team2025gemini} have demonstrated enhanced generalization capabilities by collecting diverse, large-scale datasets within controlled environments, where certain factors such as tasks, scene types, and embodiments are fixed, while others are varied. This mitigates data fragmentation, supporting our theoretical framework.
    

    
    \textbf{Shortcut learning in neural networks. } Neural networks are known to exploit spurious correlations for decision-making, leading to the shortcut learning of non-robust features or confounding factors, which can significantly hinder generalization \cite{shortcut,izmailov2022feature,ye2024spurious}. In vision tasks, neural networks have been observed to rely on multiple task-irrelevant factors, including image backgrounds \cite{xiao2020noise,sagawa2019distributionally,luo2021rectifying,moayeri2022comprehensive}, secondary objects \cite{kolesnikov2016improving,rosenfeld2018elephant,singla2021salient,shetty2019not,alcorn2019strike,mo2021object},  object textures \cite{geirhos2018imagenet} and other confounding factors \cite{brendel2019approximating,li2018resound}. In the language domain, recent studies have demonstrated that large language models tend to exploit dataset biases as shortcuts for making predictions in various downstream tasks \cite{ribeiro2020beyond,yuan2024llms,tang2023large,du2023shortcut}. While there are a few works discussing shortcut learning in reinforcement learning \cite{ding2023seeing,grandien2024interpretable,hoftijzer2023language,deng2023causal,tian2023matters} and imitation learning \cite{sanchez2024recon,de2019causal,park2021object,bica2021invariant,chen2025improving}, to the best of our knowledge, we are the first to investigate shortcut learning in generalist robot policies developed through imitation learning on large-scale datasets. Building on this work, recent studies have proposed new methods to mitigate shortcut learning in gereralist robot policies, either by modifying the training data \cite{zhang2025inspire} or through training-free approaches \cite{wu2025policy}.

    Recent works have also applied information-theoretic concepts to robotics. For instance, Hejna et al. \cite{hejna2025robot} use mutual information estimators to score the quality of individual demonstration trajectories for data curation, focusing on intra-trajectory properties like action diversity and predictability. Separately, Bai et al. \cite{bai2025rethinking} apply the Information Bottleneck principle as a regularization technique during training to mitigate redundancy in the model's latent representations. Our work is distinct from both. Rather than using mutual information as a trajectory scoring function or a model regularizer, we employ it as a diagnostic tool at the dataset-structure level. Our analysis reveals how properties between sub-datasets—namely fragmentation and limited diversity—give rise to spurious correlations. We demonstrate that these structural flaws are a root cause of shortcut learning, a problem orthogonal to the quality of individual trajectories or the redundancy of a model's learned representation.

    Recent research has increasingly recognized the critical role of data quality in imitation learning, moving beyond simple heuristics like dataset size. For instance, Belkhale et al. \cite{belkhale2023data} provide a formalism for data quality through the lens of distribution shift, identifying key intra-trajectory properties like action divergence and transition diversity as crucial for policy performance. Operating at the level of entire datasets, Hejna et al. \cite{hejna2024re} tackle the challenge of composing large-scale, heterogeneous data mixtures. Their method, Re-Mix, uses distributionally robust optimization to learn optimal sampling weights for different data domains, demonstrating that the composition of the training data has an outsized impact on the final policy's generalization capabilities. While these works aim to mitigate distribution shift by analyzing trajectory-level properties or by optimizing the dataset mixture, our work addresses the distinct but related problem of \emph{shortcut learning}. Our contribution is a novel analysis at the dataset-structure level. We use information-theoretic principles not as a trajectory scoring function or a mixture optimization objective, but as a diagnostic tool to reveal how structural flaws---namely high fragmentation and low diversity across sub-datasets---are a fundamental cause of the spurious correlations that lead to shortcut behaviors. Thus, our focus is on diagnosing the origin of a specific failure mode rooted in the dataset's structure, rather than on general data curation or mixture optimization.
    
    A recent study \cite{gao2024efficient} also investigates generalization across factors in a compositional manner, similar to the setting we study in Figure \ref{fig:intro}. However, the primary focus of that work is on optimizing data collection to cover all possible factors, rather than investigating shortcuts or spurious correlations.

\begin{figure*}[t]
      \centering
      \includegraphics[width = 0.5\linewidth]{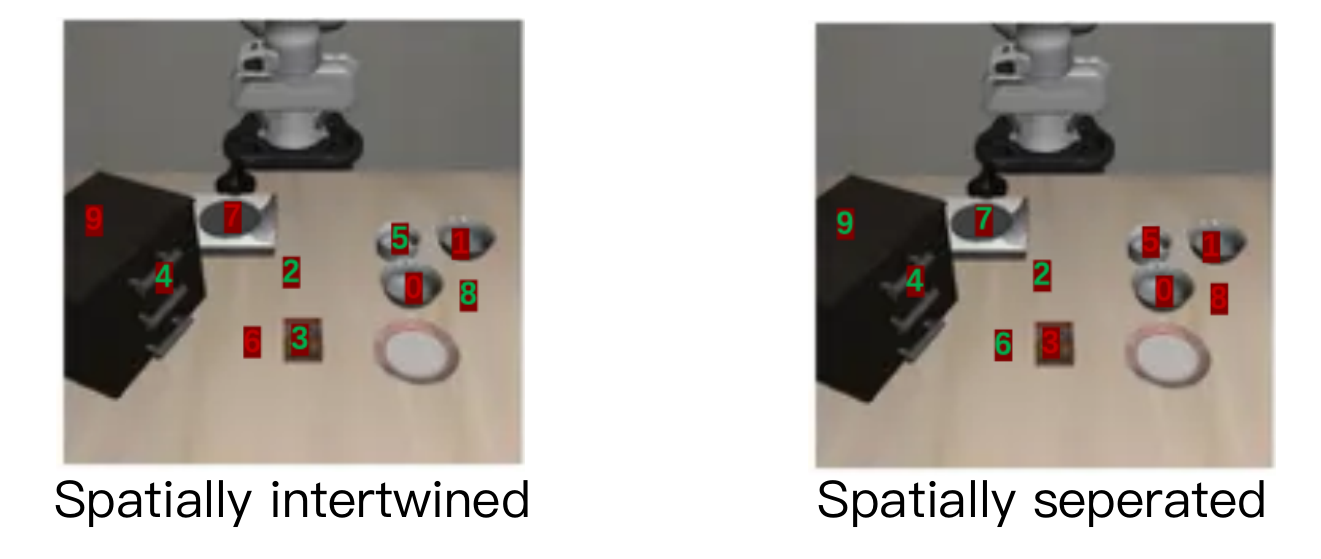}
      \caption{\textbf{Experimental setup for object position layouts across 10 tasks.} Objects from the same sub-dataset share the same color. In the left plot, object positions are spatially intertwined between sub-datasets, whereas in the right plot, they are spatially separated. Unless otherwise specified, experiments employ the high-disparity configuration (right).}
      \label{fig:Disparity_of_ten_subdataset}
\end{figure*}

\section{The Influence of Temperature}
\label{sec:Appendix_temperature}

\begin{figure*}[t]
      \centering
      \includegraphics[width = 0.7\linewidth]{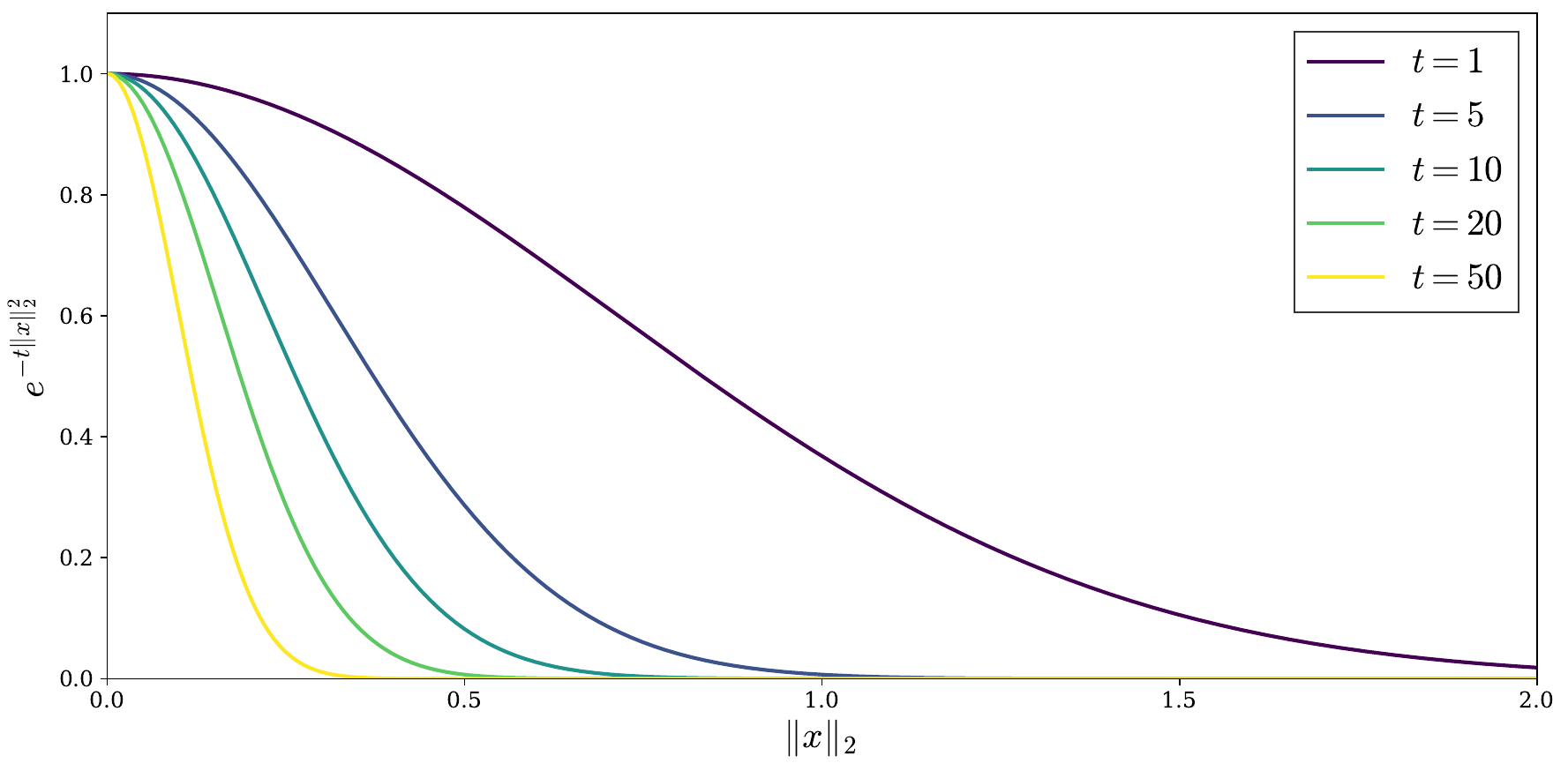}
      \caption{The similarity metric in \cite{wang2020understanding} when varying the temperature $t$. }
      \label{fig:metric_function}
\end{figure*}

In Figure \ref{fig:metric_function}, we present a visualization of the similarity metric function $e^{-t\|x\|^2_2}$, as discussed in Section \ref{sec:Dataset_Analysis}. This function is examined under varying values of the temperature parameter $t$. As $t$ increases, the function's value approaches zero more rapidly. Consequently, the temperature $t$ effectively establishes a soft threshold, which governs the range of $\|x\|_2$ over which the function maintains a value greater than zero.

\section{Additional Dataset Analysis}
\label{sec:Appendix_Dataset_Analysis}

\begin{figure*}[t]
      \centering
      \includegraphics[width = 1.0\linewidth]{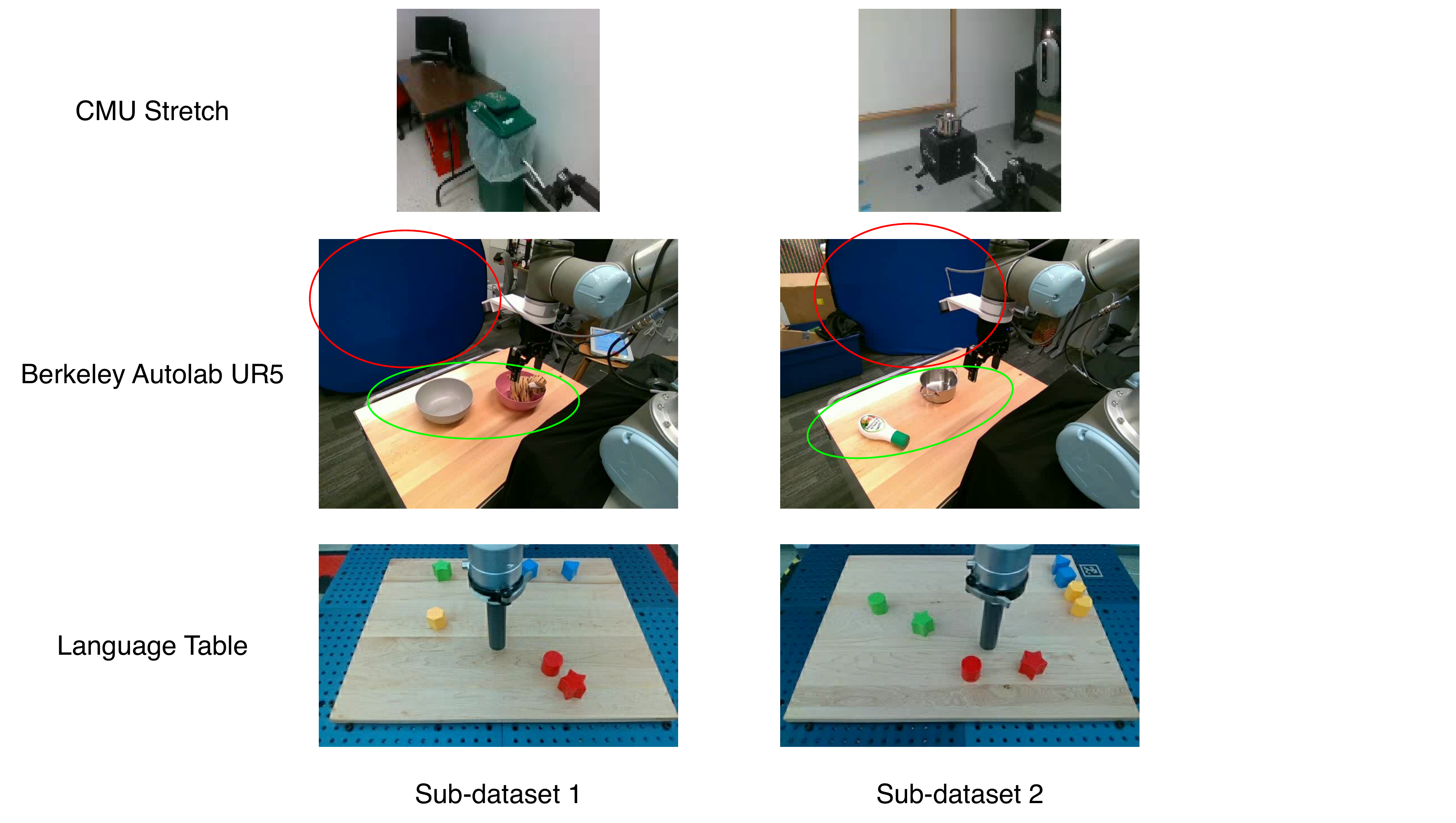}
      \caption{Three fragmented sub-datasets from OXE, each demonstrating distinct fragmentation patterns: (1) CMU Stretch, decomposable into disjoint scenes and tasks; (2) Berkeley Autolab UR5, featuring several factor  with time-correlated variations (e.g., background and tasks); (3) Language Table, with only one sparsely changing factor (e.g., lighting).}
      \label{fig:subdataset_fragmentation}
\end{figure*}

\begin{paragraph}{Sub-Dataset Fragmentation Analysis}
Figure~\ref{fig:subdataset_fragmentation} illustrates three characteristic fragmentation patterns of sub-datasets in OXE: (1) \textit{Language Table} exhibits natural clustering due to infrequent lighting changes, creating factor-independent subsets without inducing shortcut learning; (2) \textit{Berkeley AutoLab UR5} demonstrates unintended time-correlated variations where task segments coincide with background changes from human activity, creating spurious task-background correlations that promote shortcut learning; (3) \textit{CMU Stretch} (OXE sub-datasets) contains disjoint scenes with simultaneously varying environmental factors and tasks, forming strongly correlated subsets that exacerbate shortcut learning. These patterns highlight how different data collection processes of each sub-dataset can inadvertently create problematic correlations between environmental factors and tasks, aligning with our experiment results in Figure \ref{fig:10_subdatasets}.
\end{paragraph}

\begin{figure*}[t]
      \centering
      \includegraphics[width = 1.0\linewidth]{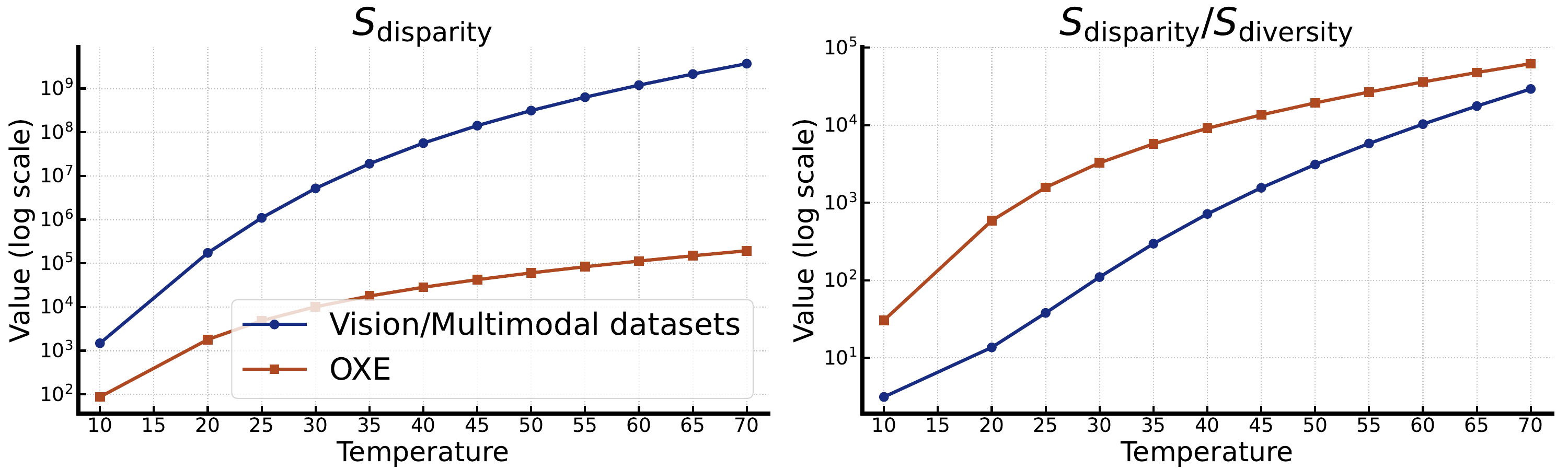}
      \caption{Comparison of the textual disparity metric $S_{\mathrm{disparity}}$ (left) and the combined metric $\frac{S_\mathrm{disparity}}{S_\mathrm{diversity}}$ (right) between OXE and vision/multimodal datasets at different temperatures.}
      \label{fig:text_disparity}
\end{figure*}

\begin{figure*}[t]
      \centering
      \includegraphics[width = 1.0\linewidth]{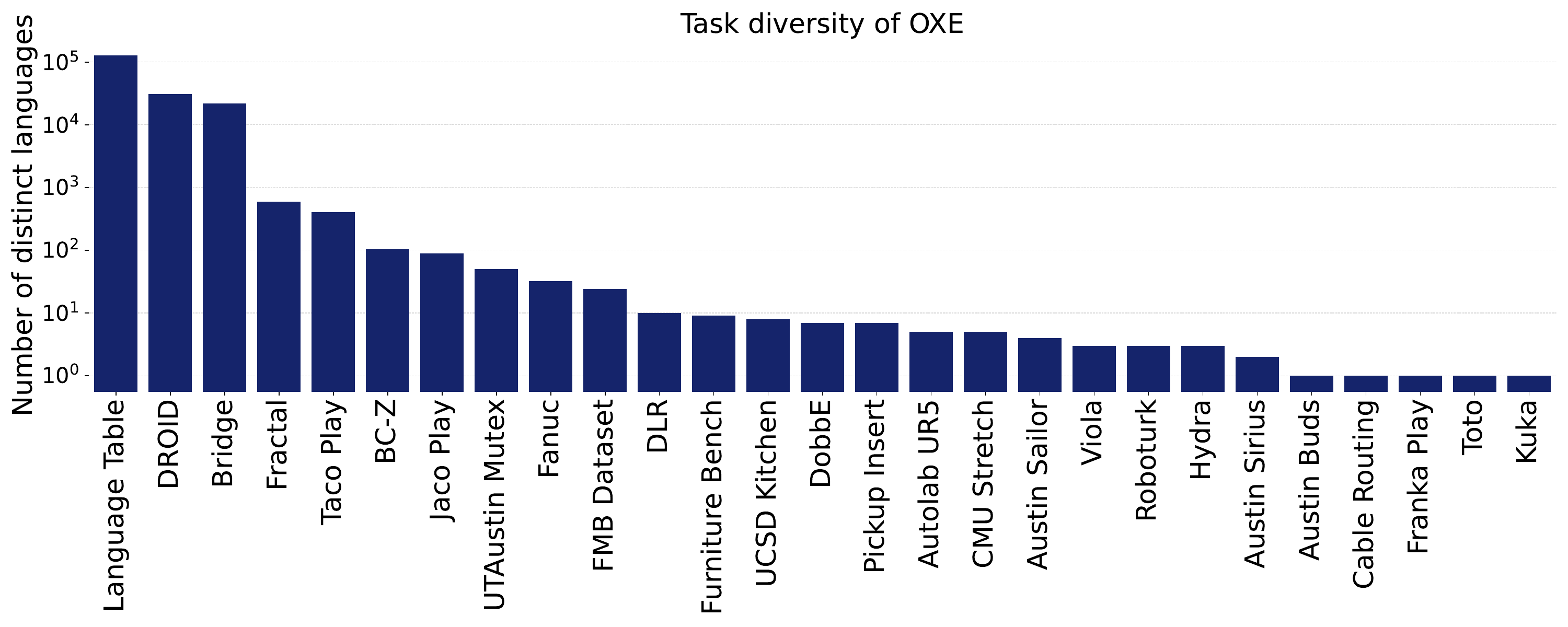}
      \caption{The number of distinct tasks (languages) within each sub-dataset of OXE. Most sub-datasets only have less than 10 tasks, which leads to extremely low task diversity.}
      \label{fig:oxe_task_diversity}
\end{figure*}

\begin{table}[t]

\centering
\caption{Sub-dataset task overlap statistics of OXE}
\label{tab:Text framentation}
\begin{tabular}{@{}lr@{}}
\toprule
\multicolumn{1}{c}{\textbf{Metric}} & \multicolumn{1}{c}{\textbf{Value}} \\
\midrule
Total tasks & \num{182158} \\
Total overlapping tasks between datasets & \num{165} \\
Percentage of overlapping sub-dataset pairs  & \SI{3.70}{\percent} \\

\bottomrule
\end{tabular}
\end{table}

%

\section{LIBERO Experiment Details}

\subsection{Model Implementation and Training}
For our analysis, we implemented and trained three distinct models, each with specific configurations:
\begin{itemize}
    \item \textbf{Diffusion Policy:} This model uses a ResNet-18 vision backbone with images resized to $84 \times 84$. It was trained for 30,000 iterations (batch size 128) using an AdamW optimizer (learning rate $1 \times 10^{-4}$, weight decay $1 \times 10^{-6}$). The model takes a 2-step observation history as input, excluding proprioception. Training required approximately 5 hours on a single NVIDIA 3090 GPU.

    \item \textbf{MiniVLA:} This vision-language-action model uses Vector-Quantized action chunks (horizon=8) and was trained without wrist camera images, proprioception, or historical state data. It was optimized for 10,000 steps (batch size 128) with a constant learning rate of $2 \times 10^{-5}$. Training was distributed across eight NVIDIA A6000 GPUs and took 5 hours.

    \item \textbf{$\pi_0$:} This model integrates a PaliGemma 2B backbone (using LoRA) with a 300M-parameter action expert (action dimension 32, horizon 50). It was trained for 10,000 steps (batch size 32) using AdamW with a cosine decay learning rate schedule (1,000-step warmup to a peak LR of $2.5 \times 10^{-5}$, decaying to $2.5 \times 10^{-6}$ over 30,000 steps). Training took 8 hours across four NVIDIA A6000 GPUs.
\end{itemize}

\subsection{Experimental Environment and Task Setup}
Our evaluations were conducted within the \textbf{LIBERO-Spatial} suite. The fundamental goal for all 10 manipulation tasks is to \textbf{place the target bowl into the red plate}.
\begin{itemize}
    \item \textbf{Task-Relevant Factors}: The 10 distinct tasks are defined by the \textbf{initial position of the target bowl} (e.g., in a drawer, on a shelf). The corresponding \textbf{language instruction} changes accordingly to reflect this initial position (e.g., "\textit{pick up the bowl in the top drawer and place it on the red plate}").
    \item \textbf{Task-Irrelevant Factor}: We focused on the \textbf{camera viewpoint}, defined by $\theta \in [-10^\circ, 90^\circ]$, as the primary task-irrelevant factor.
    \item \textbf{Scene Simplification}: The original LIBERO environment contains two bowls. To better isolate the factors of interest, we \textbf{removed one bowl}, leaving only a single target object (marked in red in Figure 5). This simplification also accommodates vision-only models like Diffusion Policy.
\end{itemize}

\subsection{Data Collection}
For each experimental setting shown in Figure 6, we used \textbf{200 demonstrations for each task}. These were generated by sampling 4 random viewpoints for each of the 50 base trajectories provided by LIBERO for that task.

\subsection{Protocol for Viewpoint Diversity and Disparity Experiments}
\label{sec:Experiment_Details}
\begin{itemize}
    \item \textbf{Parameter Selection Strategy}: For both the diversity and disparity experiments, the specific viewpoint centers and radii were not chosen arbitrarily. They were \textbf{systematically selected to identify the critical range where the policy's behavior transitions from robust to shortcut-reliant}. This allowed us to precisely map out the model's sensitivity to these dataset properties.
    \item \textbf{Diversity Protocol (Fig. 6 Left)}: We systematically increased the \textit{range} (radius) of viewpoints for each task during training, while holding the \textit{centers} of the viewpoint distributions constant. Evaluation was performed at fixed, out-of-distribution viewpoints to fairly assess generalization.
    \item \textbf{Disparity Protocol (Fig. 6 Middle)}: Conversely, we varied the \textit{distance between the centers} of the viewpoint distributions while keeping their \textit{radius} constant and narrow. To ensure a challenging test, evaluation points were always selected from the boundaries of the opposing task's distribution.
\end{itemize}

\section{Real-world Experiment Setup}
\label{Real-world Experiment Setup}
This section details the setup for the two distinct real-world experiments presented in the paper. Both experiments utilize an AgileX PIPER robotic arm and are observed by two cameras from different viewpoints.

\subsection{Experiments in Figure 1}
This experiment is designed to test if a model exhibits shortcut learning when task-irrelevant factors are confounded, even when all objects are present during training.

\begin{itemize}
    \item \textbf{Task-Relevant Factors}: The identity of the target object (banana or watermelon) and the corresponding language instruction ("place tissue bag into the plate" or "place snack bag into the plate").
    \item \textbf{Task-Irrelevant Factor}: The camera viewpoint (left or right camera).
    \item \textbf{Training Data Setup}: Two sub-datasets were created.
        \begin{itemize}
            \item \textbf{Sub-dataset 1}: The instruction is "place tissue bag into the plate", collected exclusively from the \textbf{left camera}.
            \item \textbf{Sub-dataset 2}: The instruction is "place snack bag into the plate", collected exclusively from the \textbf{right camera}.
        \end{itemize}
    \item \textbf{Data Collection Details}: A key difference from the object augmentation experiment is that during the collection of each demonstration, \textbf{both the tissue bag and the snack bag were present on the table}. The only sources of randomness were the minor variations in the orientation of the objects and the slight shifts in their positions and the position of the plate. We collected 20 demonstrations for each sub-dataset.
    \item \textbf{Evaluation}: The model is evaluated on its ability to follow the correct instruction when the viewpoint is swapped (e.g., given the "place snack bag..." instruction from the left camera's viewpoint). The model's ability to follow the instruction was measured over 10 trials for each condition.
\end{itemize}

\begin{figure*}[t]
      \centering
      \includegraphics[width = 0.9\linewidth]{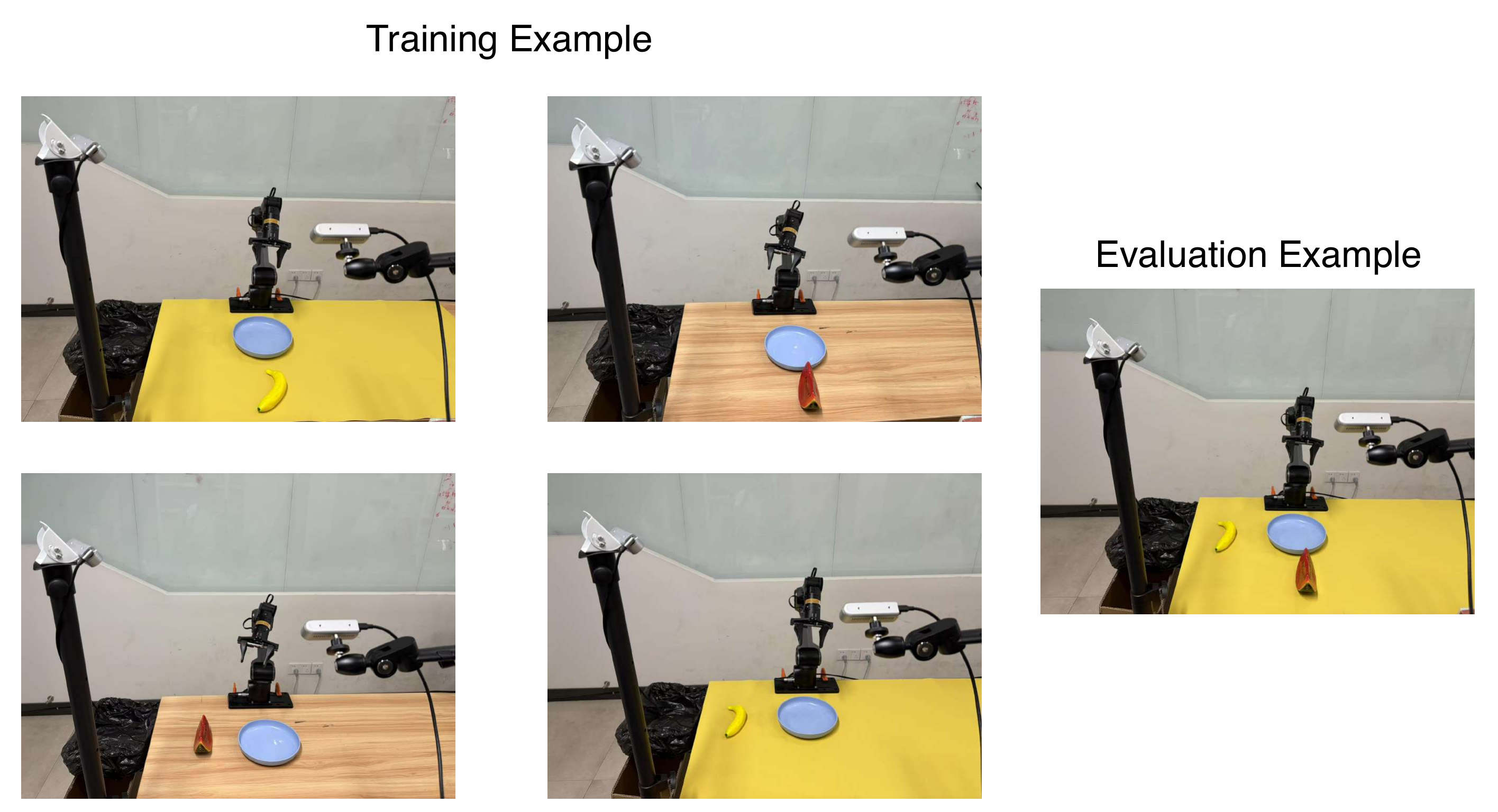}
      \caption{\textbf{Real-world object augmentation experiment setup.} Two cameras are positioned in front of an AgileX PIPER robotic arm. For training, two distinct sub-datasets were created, each featuring a single, specific combination of object type (banana or watermelon), camera viewpoint (left or right), and background (with or without a yellow tablecloth). During evaluation, tests were conducted on both two original combinations of viewpoints and backgrounds. In these evaluations, both objects (banana and watermelon) were simultaneously present on the table, and the robot was guided by language instructions referring to object-scene configurations \emph{not} explicitly encountered in the training combinations.}

      \label{fig:real_world_setup}
    \vspace{-0.5cm}
    \end{figure*}

\subsection{Object Augmentation Experiments}
This experiment is designed to create a strong spurious correlation between objects and multiple visual factors (viewpoint and background) and to test if data augmentation can mitigate the resulting shortcut behavior.

\begin{itemize}
    \item \textbf{Task-Relevant Factors}: The identity of the target object (banana or watermelon) and the corresponding language instruction.
    \item \textbf{Task-Irrelevant Factors}: The camera viewpoint, the background scene, and the positions of objects.
    \item \textbf{Training Data Setup (Confounded)}: These irrelevant factors were deliberately confounded with the task-relevant object.
        \begin{itemize}
            \item \textbf{Sub-dataset 1 ("Banana Env")}: The task "put banana into the plate" was collected exclusively from the \textbf{left camera} with a \textbf{yellow tablecloth} background.
            \item \textbf{Sub-dataset 2 ("Watermelon Env")}: The task "put watermelon into the plate" was collected exclusively from the \textbf{right camera} with \textbf{no tablecloth}.
        \end{itemize}
    \item \textbf{Data Collection Details}: During training data collection for each sub-dataset, only the single relevant object (either the banana or the watermelon) was present. The object was randomly placed either on the plate or to its right. Each sub-dataset consists of 20 demonstrations.
    \item \textbf{Evaluation}: To test for shortcut learning, the learned correlations were disrupted by introducing out-of-distribution (OOD) objects into the scene (e.g., presenting the banana in the right-camera, no-tablecloth environment) and providing the corresponding language instruction. The model's ability to follow the instruction was measured over 10 trials for each condition.
\end{itemize}

\section{Object Augmentation Details}
\label{appendix: data_aug}

\subsection{Data Collection and Training Setup}
For our experiments involving object augmentation, we established the following datasets and training protocols:
\begin{itemize}
    \item \textbf{SIMPLER Environment:} We collected a total of 242 successful trajectories, comprising 116 from the RT-1 environment and 126 from the Bridge environment.
    \item \textbf{Real-World Environment:} We collected 20 demonstrations for each of the two distinct sub-datasets.
    \item \textbf{Training Protocol:} All models, whether using original or augmented data, were fine-tuned for 2,500 steps to ensure a fair comparison.
\end{itemize}

\subsection{Augmentation Pipeline}
Our object augmentation pipeline is designed to decouple objects from their original visual contexts. It consists of the following three stages:

\begin{enumerate}
    \item \textbf{Step 1: Object Mask Library Creation.}
    First, we build a comprehensive object mask library ($D$). To do this, we apply Grounded-SAM2 to the initial frame of every episode in our dataset. This allows us to extract high-quality segmentation masks for all target objects as seen from various perspectives, creating a rich library of object assets.

    \item \textbf{Step 2: Scene Preparation (Object Removal and Inpainting).}
    Next, for each image ($o_t$) in a trajectory, we identify the target object specified by the language instruction ($L$). Using Grounded-SAM2 again, we segment this specific object to obtain its mask ($m_{orig}$) and record its original position (via the bounding box center, $c_{orig}$). The object is then digitally erased from the image, and an inpainting model seamlessly reconstructs the background where the object was. This step yields a "clean" background image, ready for augmentation.

    \item \textbf{Step 3: Object Swapping and Augmented Dataset Generation.}
    Finally, we create the augmented image. For a clean scene that originally contained a specific object (e.g., a banana with mask $d_{orig}$), we randomly sample a mask of a \textit{different} object from our library (e.g., a watermelon, $d_{aug} \in D \setminus \{d_{orig}\}$). We then paste this new object into the clean scene, carefully aligning its center with the original object's location ($c_{orig}$). To create a more challenging OOD scenario, we then re-introduce the original object into the same scene to act as a distractor. The placement of this distractor depends on the environment: in SIMPLER, it is placed at a random valid location, while in our real-world setup, it is placed in one of the two predefined object locations. By applying this full procedure—swapping the target and adding a distractor—across all images, we generate the final augmented dataset where objects are fully decoupled from their contexts.
\end{enumerate}

\section{Methodology for Human-Assisted Shortcut Scoring}

To quantitatively measure the degree to which a policy relies on shortcut learning during out-of-distribution (OOD) evaluations, we developed a human-assisted scoring methodology. This approach allows for a nuanced assessment of the robot's behavior beyond simple binary success/failure metrics.

The scoring process is conducted as follows:
\begin{enumerate}
    \item \textbf{Video Review:} Human evaluators are presented with video recordings of every evaluation trial for a given experimental setup. Each video captures the complete sequence of actions taken by the robot policy from the start to the end of an episode.

    \item \textbf{Behavioral Judgment:} For each video, the evaluator judges whether the policy's actions correspond to the given language instruction or if they revert to a ``shortcut'' behavior learned from spurious correlations in the training data.

    \item \textbf{Scoring Rubric:} A score is assigned to each trial based on a predefined rubric:
    \begin{itemize}
        \item \textbf{Score = 1.0 (Clear Shortcut):} The policy unequivocally ignores the instruction and performs a clear shortcut action. For example, when instructed to interact with object A from a viewpoint previously associated with object B, the policy ignores A and attempts to interact with B.
        \item \textbf{Score = 0.5 (Ambiguous or Partial Shortcut):} The policy's behavior is unclear or appears to be a mix of the correct and shortcut actions. This includes cases where the robot targets a location midway between the correct object (A) and the shortcut object (B), or exhibits significant hesitation.
        \item \textbf{Score = 0.0 (No Shortcut):} The policy correctly attempts to follow the language instruction and does not exhibit any observable shortcut behavior, regardless of whether the attempt is successful or not.
    \end{itemize}

    \item \textbf{Final Score Calculation:} The final ``Degree of Shortcut Learning'' for a model is calculated by averaging the scores from all of its evaluation trials. A score closer to 1.0 indicates a strong tendency to rely on shortcuts, while a score closer to 0.0 indicates that the policy is more robust to the spurious correlations present in the training data.
\end{enumerate}

\section{Proofs of Propositions in Section \ref{sec:Theoretical_and_Empirical_Analysis}}
\label{sec:Appendix_proof}
\textbf{Proof of Proposition \ref{prop:diversity}}:
\begin{proof}
Given the condition $p_u(u)=\frac{1}{2}\left[p_{u_1}(u)+p_{u_2}(u)\right]$ and $H(u)=-\sum_{u\in U}p_u(u)\log p_u(u)$, we have
\begin{align*}
H(u)&=-\sum_{u\in U_1}\frac{p_{u_1}(u)}{2}\log \frac{p_{u_1}(u)}{2}-\sum_{u\in U_2}\frac{p_{u_2}(u)}{2}\log \frac{p_{u_2}(u)}{2}\notag\\
&=\frac{H(u_1)+H(u_2)}{2}+1.
\end{align*}
The last equation comes from the fact that $\sum_{u\in U_1}p_{u_1}(u)\log\frac{p_{u_1}(u)}{2}=H(u_1)+\log2$ and $log_22=1$.
Similarly, $H(v)=\frac{H(v_1)+H(v_2)}{2}+1.$ 
For the mutual information, since the assumption of independence of factors within each sub-dataset, we have $I(u_1,v_1)=I(u_2,v_2)=0$, and thus
\begin{align*}
I(u, v) &= \sum_{u\in U_1, v\in V_1} \frac{p_1(u, v)}{2} \log \frac{\frac{p_1(u, v)}{2}}{\frac{p_{u_1}(u)}{2} \cdot \frac{p_{v_1}(v)}{2}} + \sum_{u\in U_2, v\in V_2} \frac{p_2(u, v)}{2} \log \frac{\frac{p_2(u, v)}{2}}{\frac{p_{u_2}(u)}{2} \cdot \frac{p_y(v)}{2}}\notag\\
&= \frac{I(u_1,v_1)+I(u_2,v_2)}{2}+1\notag\\
&= 1.
\end{align*}
Put together $H(u)$, $H(v)$ and $I(u,v)$, we have
\begin{align*}
    \overline{I}(u,v) &= \frac{2I(u,v)}{H(u)+H(v)}\notag\\
    &=\frac{4}{H(u_1)+H(u_2)+H(v_1)+H(v_2)+4}\notag\\
    &=\frac{4}{C_{\mathrm{diversity}}+4},
\end{align*}
which completes the proof.
\end{proof}

\textbf{Proof of Proposition \ref{prop:fragmentation}:}
\begin{proof}
Since both sub-datasets involve probabilities over \( U_{12} \) and \( V_{12} \), we should consider each region separately. First, we calculate the entropy \( H(u) \):
\begin{align*}
    H(u) =& - \sum_{u\in U_1\backslash U_{12}} \frac{p_{u_1}(u)}{2} \log \frac{p_{u_1}(u)}{2} - \sum_{u\in U_2\backslash U_{12}} \frac{p_{u_2}(u)}{2} \log \frac{p_{u_2}(u)}{2}\notag\\
    &- \sum_{u\in U_{12}} \frac{p_{u_1}(u) + p_{u_2}(u)}{2} \log \frac{p_{u_1}(u) + p_{u_2}(u)}{2}.
\end{align*}
By applying Jensen Inequality, we have 
\begin{equation}
    \frac{p_{u_1}(u) + p_{u_2}(u)}{2} \log \frac{p_{u_1}(u) + p_{u_2}(u)}{2}\leq \frac{p_{u_1}(u)}{2}\log p_{u_1}(u)+\frac{p_{u_2}(u)}{2}\log p_{u_2}(u),\notag
\end{equation}
which gives
\begin{align*}
    H(u) \geq&  - \sum_{u\in U_1\backslash U_{12}} \frac{p_{u_1}(u)}{2} \log \frac{p_{u_1}(u)}{2} - \sum_{u\in U_2\backslash U_{12}} \frac{p_{u_2}(u)}{2} \log \frac{p_{u_2}(u)}{2}\notag\\
    &-\sum_{u\in U_{12}}\left[\frac{p_{u_1}(u)}{2}\log p_{u_1}(u)+\frac{p_{u_2}(u)}{2}\log p_{u_2}(u)\right]\notag\\
    =&\frac{1}{2}\left[H(u_1)+H(u_2)+\sum_{u\in U_1\backslash U_{12}}p(u_1)+\sum_{u\in U_2\backslash U_{12}}p(u_2)\right]\notag\\
    =&\frac{1}{2}\left[H(u_1)+H(u_2)+2-\sum_{u\in U_{12}}p(u_1)-\sum_{u\in U_{12}}p(u_2)\right].
\end{align*}
Similarly, for $H(v)$, we have
\begin{align*}
H(v) \geq&\frac{1}{2}\left[H(v_1)+H(v_2)+2-\sum_{v\in V_{12}}p(v_1)-\sum_{v\in V_{12}}p(v_2)\right].
\end{align*}
Given that $C_{\mathrm{interleave}}=\sum_{u\in U_{12}}\left[p_{u_1}(u)+p_{u_2}(u)\right] + \sum_{v\in V_{12}} \left[p_{v_1}(v)+p_{v_2}(v)\right]$, we have
\begin{align*}
    H(u)+H(v)\geq \frac{1}{2}\left[C_{\mathrm{diversity}}+4-C_{\mathrm{interleave}})\right].
\end{align*}
Then we calculate the mutual information. We partition the calculation into four terms:
\begin{align*}
I(u,v)=&\sum_{i=1}^2\bigg(\sum_{u\in U_i\backslash U_{12},v\in V_i\backslash V_{12}}\frac{p_i(u,v)}{2}\log\frac{2p_i(u,v)}{p_{u_i}(u)p_i(v)}\bigg)\notag\\
&+\sum_{u\in U_{12},v\in V}p(u,v)\log\frac{p(u,v)}{p_u(u)p_v(v)}+\sum_{u\in U,v\in V_{12}}p(u,v)\log\frac{p(u,v)}{p_u(u)p_v(v)}.
\end{align*}
We first calculate the last two terms. Note that
\begin{align*}
    &\sum_{u\in U_{12},v\in V_1\cup V_{12}}p(u,v)\log\frac{p(u,v)}{p_u(u)p_v(v)}\notag\\
    &=\sum_{u\in U_{12},v\in V_1\cup V_{12}}\frac{1}{4}\left[p_1(u,v)(1+\frac{p_{u_2}(u)}{p_{u_1}(u)})\right]\log\frac{p_1(u,v)}{p_{u_1}(u)p_{v_1}(v)}.\notag\\
\end{align*}
Since $u$ and $v$ are independent in the first sub-dataset, we have $p_1(u,v)=p_{u_1}(u)p_{v_1}(v)$, and thus 
\begin{align*}
    \sum_{u\in U_{12},v\in V_1\cup V_{12}}p(u,v)\log\frac{p(u,v)}{p_u(u)p_v(v)}=0.
\end{align*}
Similarly, we have

\begin{align*}
\sum_{u\in U_{12},v\in V_2}p(u,v)\log\frac{p(u,v)}{p_u(u)p_v(v)}=0,
\end{align*}
and thus 
\begin{align*}
    \sum_{u\in U_{12},v\in V}p(u,v)\log\frac{p(u,v)}{p_u(u)p_v(v)}=0,
\end{align*}
and similarly,
\begin{align*}
    \sum_{u\in V_{12},u\in U}p(u,v)\log\frac{p(u,v)}{p_u(u)p_v(v)}=0.
\end{align*}
Thus, we only need to calculate the first two terms:
\begin{align*}
I(u,v)=&\sum_{i=1}^2\sum_{u\in U_i\backslash U_{12},v\in V_i\backslash V_{12}}\frac{p_i(u,v)}{2}\log\frac{2p_i(u,v)}{p_{u_i}(u)p_i(v)}\notag\\
=&\sum_{u\in U_1\backslash U_{12},v\in V_1\backslash V_{12}}\frac{p_1(u,v)}{2}+\sum_{u\in U_2\backslash U_{12},v\in V_2\backslash V_{12}}\frac{p_2(u,v)}{2}\notag\\
=&\frac{1}{2}\left[\sum_{u\in U_1\backslash U_{12}}p_{u_1}(u)\sum_{v\in V_1\backslash V_{12}}p_{v_1}(v)+\sum_{u\in U_2\backslash U_{12}}p_{u_2}(u)\sum_{V\in V_2\backslash V_{12}}p_{v_2}(v)\right]\notag\\
\leq &\frac{1}{4}\left[\sum_{u\in U_1\backslash U_{12}}p_{u_1}(u)+\sum_{v\in V_1\backslash V_{12}}p_{v_1}(v)+\sum_{u\in U_2\backslash U_{12}}p_{u_2}(u)+\sum_{V\in V_2\backslash V_{12}}p_{v_2}(v)\right]\notag\\
=&\frac{1}{4}\left(4-C_{\mathrm{interleave}}\right).
\end{align*}

Put together, we have
\begin{align*}
\overline{I}(u,v)\leq &\frac{4-C_{\mathrm{interleave}}}{C_{\mathrm{diversity}}+4-C_{\mathrm{interleave}}}\notag\\
=&1-\frac{C_{\mathrm{diversity}}}{C_{\mathrm{diversity}}+\left(4-C_{\mathrm{interleave}}\right)},
\end{align*}

which completes the proof.
\end{proof}

\section{Linear Model Analysis for the Impact of Disparity Between Sub-datasets on Shortcut Learning}
\label{sec:linear_model_analysis}
We consider a simple linear model defined as $\pi_\theta(x) = \pi_\theta([u,v]) = \omega^T[u,v] + b = \omega_1^T u + \omega_2^T v + b$, where the factor generation model $g$ is assumed to be the identity map. We further assume that the sum of prediction errors is zero, i.e., $\mathbb{E} \left[y - \pi_\theta(x)\right] = 0$. This condition can be satisfied by adjusting the bias term $b$ to $\mathbb{E} \left[y - \omega^T[u,v]\right]$. Our focus is on the gradient descent optimization of the parameter $\omega$ using the $L_2$ loss function. The gradients with respect to $\omega_1$ and $\omega_2$ are given by:
\[
[g_{\omega_1}, g_{\omega_2}] = \left[-2\mathbb{E}(y - \omega_1^T u - \omega_2^T v - b)u, -2\mathbb{E}(y - \omega_1^T u - \omega_2^T v - b)v\right].
\]
The magnitudes of these gradients determine the relative importance of the factors $u$ and $v$ in the model's decision-making process. For simplicity, we assume $\mathbb{E} u = \mathbb{E} v = 0$, which does not affect the gradients (by setting $u\leftarrow u-\mathbb{E}u$). Assuming an initial weight of zero, the initial gradient is:
\[
[g_{\omega_1}, g_{\omega_2}] = \left[-2\mathbb{E}((y - \mathbb{E}y)(u - \mathbb{E}u)), -2\mathbb{E}((y - \mathbb{E}y)(v - \mathbb{E}v))\right].
\]
This expression reveals that the gradients measure the correlations between the factors $u, v$ and the target variable $y$. Importantly, these correlations are strongly influenced by the scale of $u - \mathbb{E}u$ and $v - \mathbb{E}v$. The distance between the factors of sub-datasets significantly affects these scales. Consider a scenario where the distance $d(U_1, U_2)$ is increased to $t \cdot d(U_1, U_2)$ without altering the content of $u_1$ and $u_2$. In this case, $\mathbb{E}((y - \mathbb{E}y)(u - \mathbb{E}u))$ will approximately increase to $t \cdot \mathbb{E}((y - \mathbb{E}y)(u - \mathbb{E}u))$, as the increased distance increases the scale and variance of the random variable $u$ by the same extent.  Thus, the distances between factors of sub-datasets play a crucial role in determining whether shortcut learning occurs. If spurious correlations exist and the sub-dataset distance of task-irrelevant factors $d(V_1, V_2)$ is significantly greater than that of task-relevant factors $d(U_1, U_2)$, the model is more likely to learn shortcuts.

\section{Vision and Multimodal Datasets}
\label{sec:Appendix_dataset_introduction}
We list the vision and multimodal datasets we use in Section \ref{sec:Dataset_Analysis}:

\textbf{ImageNet-1K} \cite{ImageNet}: ImageNet is a large-scale visual database designed for use in visual tasks. It contains over 14 million images that have been hand-annotated to indicate what objects are picture, and ImageNet-1K is a subset that contains more than 1M images with one thousand classes. It has been widely used for training large-scale vision models, including recent self-supervised models that have been used as the visual encoder for vision-language models.

\textbf{Open Images} \cite{Open_images}: Open Images is a large-scale dataset for object detection, segmentation, and visual relationship detection, containing over 9 million images annotated with image-level labels, object bounding boxes, and visual relationships. It provides a comprehensive resource for developing and benchmarking models in various computer vision tasks, with a focus on real-world image diversity and complexity. We use the sixth version of the dataset.

\textbf{COCO} \cite{COCO}: The Common Objects in Context (COCO) dataset is a large-scale object detection, segmentation, and captioning dataset. It contains over 330,000 images, with more than 200,000 labeled images and 1.5 million object instances. It has often been used as part of the instruction tuning dataset for vision-language models \cite{llava}.

\textbf{ADE20K} \cite{ADE20K}: ADE20K is a dataset for semantic segmentation and scene parsing, containing over 20,000 images covering a wide range of scenes and object categories. Each image is densely annotated with objects and stuff categories.

\textbf{iNaturalist} \cite{inaturalist}: The iNaturalist dataset is a large-scale species classification dataset, derived from the iNaturalist community, which is a citizen science project and online social network of naturalists. It contains millions of images spanning thousands of species.

\textbf{Flickr30k} \cite{flickr30k}: Flickr30k is a dataset for multimodal research, consisting of 31,000 images collected from Flickr. Each image is paired with five different captions.

\textbf{GQA} \cite{GQA}: The GQA (Graph Question Answering) dataset is designed for visual question answering tasks, featuring 22 million questions about 140,000 images. The dataset has been widely used for evaluation of vision-language models, and it has also been used as the tuning dataset of some vision-language models \cite{sphinx}.

\textbf{Visual Genome} \cite{Visual_genome}: Visual Genome is a dataset that connects structured image data with language, containing over 100,000 images with region descriptions, object annotations, attributes, and relationships. It serves as a comprehensive resource for tasks involving scene understanding, object detection, and relationship modeling, facilitating research in bridging vision and language.

\textbf{LAION-400M} \cite{LAION}: LAION-400M is a large-scale dataset consisting of 400 million image-text pairs, collected from publicly available Common Crawl data. It is designed to support research in large-scale multimodal learning, providing a diverse and extensive resource for training vision-language models.

\textbf{CC3M} \cite{CC3M}: The Conceptual Captions 3M (CC3M) dataset is a large-scale image captioning dataset containing approximately 3.3 million images sourced from the web. Each image is paired with a caption that describes the visual content, offering a valuable resource for training and evaluating models in vision-language tasks.

As there may be overlaps between these datasets, we filter the duplicate data before conducting the analysis in Section \ref{sec:Dataset_Analysis}.

\end{document}